\documentclass[acmsmall,screen]{acmart}
\usepackage[utf8]{inputenc}
\usepackage{hyperref}
\usepackage{multirow}
\usepackage{subcaption}
\usepackage{xcolor}

\definecolor{plt_blue}{RGB}{31, 119, 180}
\definecolor{plt_orange}{RGB}{255, 127, 14}
\definecolor{plt_green}{RGB}{44, 160, 44}

\AtBeginDocument{%
  \providecommand\BibTeX{{%
    \normalfont B\kern-0.5em{\scshape i\kern-0.25em b}\kern-0.8em\TeX}}}

\setcopyright{acmlicensed}
\acmJournal{CSUR}
\acmYear{2024} \acmVolume{1} \acmNumber{1} \acmArticle{1} \acmMonth{1}
\acmDOI{10.1145/3691339}

\begin{document}

\title{A Survey on Stability of Learning with Limited Labelled Data and its Sensitivity to the Effects of Randomness}

\author{Branislav Pecher}
\affiliation{%
  \institution{Faculty of Information Technology, Brno University of Technology}
  %\streetaddress{Mlynske nivy 5}
  \city{Brno}
  \country{Czechia}
}
\additionalaffiliation{%
  \institution{Kempelen Institute of Intelligent Technologies}
  \city{Bratislava}
  \country{Slovakia}
}
\email{branislav.pecher@kinit.sk}
\orcid{0000-0003-0344-8620}

\author{Ivan Srba}
\affiliation{%
  \institution{Kempelen Institute of Intelligent Technologies}
  \city{Bratislava}
  \country{Slovakia}
}
\email{ivan.srba@kinit.sk}
\orcid{0000-0003-3511-5337}

\author{Maria Bielikova}
\affiliation{%
  \institution{Kempelen Institute of Intelligent Technologies}
  \city{Bratislava}
  \country{Slovakia}
}
\additionalaffiliation{%
  \institution{slovak.AI}
  \city{Bratislava}
  \country{Slovakia}
}
\email{maria.bielikova@kinit.sk}
\orcid{0000-0003-4105-3494}

\renewcommand{\shortauthors}{Pecher, et al.}

\begin{abstract}
Learning with limited labelled data, such as prompting, in-context learning, fine-tuning, meta-learning or few-shot learning, aims to effectively train a model using only a small amount of labelled samples. However, these approaches have been observed to be excessively sensitive to the effects of uncontrolled randomness caused by non-determinism in the training process. The randomness negatively affects the stability of the models, leading to large variances in results across training runs. When such sensitivity is disregarded, it can unintentionally, but unfortunately also intentionally, create an imaginary perception of research progress. Recently, this area started to attract research attention and the number of relevant studies is continuously growing. In this survey, we provide a comprehensive overview of 415 papers addressing the effects of randomness on the stability of learning with limited labelled data. We distinguish between four main tasks addressed in the papers (investigate/evaluate; determine; mitigate; benchmark/compare/report randomness effects), providing findings for each one. Furthermore, we identify and discuss seven challenges and open problems together with possible directions to facilitate further research. The ultimate goal of this survey is to emphasise the importance of this growing research area, which so far has not received an appropriate level of attention, and reveal impactful directions for future research.
\end{abstract}





\maketitle

\section{Introduction}

The approaches for \textit{learning with limited labelled data} are designed to achieve high performance in machine learning models even with few labels available~\cite{liu2021pre, song2023comprehensive_survey_of_fsl}. Under the term \textit{learning with limited labelled data}, we understand any approach that is designed to work with a lack of labels, without any constraint on how the labelled samples are distributed, i.e., whether all the samples are from a single task or are distributed across different tasks. Although similar to the notion of few-shot learning, it represents a broader scope encompassing a larger number of possible approaches (some of which are incorrectly categorised as few-shot learning in current practice).

To deal with the limited labels, these approaches utilise additional information from different sources~\cite{brown2020language, song2023comprehensive_survey_of_fsl}, such as transferring knowledge from similar tasks and datasets. In the NLP domain, \textit{prompting} and \textit{in-context learning} (also called few-shot prompting) have recently emerged. In these techniques, a large pre-trained language model is ``prompted`` to predict a label for a single test sample by presenting it with task instructions and the test sample (and concatenation of a few labelled samples when \textit{in-context learning} is used), without requiring any parameter update~\cite{liu2021pre}. In addition, it is common to use \textit{fine-tuning}, where the parameters, or their subset using parameter-efficient fine-tuning (PEFT) methods, of the pre-trained large language model are updated to optimise the model for the specific downstream task using only a few labelled samples~\cite{mosbach_stability_2021, dodge_fine-tuning_2020, chen-etal-2022-revisiting}. Finally, \textit{meta-learning} can be used, where the model is explicitly trained to quickly adapt to a new task with only a handful of examples by learning how best to learn across a large number of related tasks with few labelled samples each~\cite{agarwal_sensitivity_2021, hospedales2021meta}. 

However, a significant problem observed for these approaches is their sensitivity to the effects of uncontrolled randomness, which negatively affects their stability. Under \textit{stability} and its opposite term \textit{sensitivity}, we understand a property of a learning algorithm or a model that indicates what influence the \textit{small-scale and random changes} (or perturbations) in the input \textit{data} and \textit{parameters} have on its outputs. Such random changes are introduced by different \textit{randomness factors} that represent the non-deterministic decisions in the training process, such as random initialisation of model parameters or data shuffling~\cite{pham_problems_2021, gundersen2022sources_of_irreproducibility}. These \textit{randomness factors} represent the main point around which the effects of randomness are addressed.

The uncontrolled effects of randomness can have a massive impact on the stability of the utilised machine learning approaches. Running the training multiple times on the same dataset, with the same setup and hyperparameters, may lead to large deviations in the final performance~\cite{mosbach_stability_2021, dodge_fine-tuning_2020, lu_fantastically_2022, agarwal_sensitivity_2021}. Changing only the order of samples or answers in multi-choice question answering when using in-context learning can lead the model from state-of-the-art predictions to random guesses~\cite{lu_fantastically_2022, zong2024foolvisionandlanguage, wei2024unveiling}. Choosing a different set of adaptation data in meta-learning can lead to a difference between minimum and maximum performance being up to 90\% \cite{agarwal_sensitivity_2021}. Even though the effects of randomness are present in all cases of machine learning, their impact on stability is especially significant in low-data regimes.

\begin{figure}[tbh]
    \centering
    \begin{subfigure}[b]{0.49\textwidth}
        \centering
        \includegraphics[width=\textwidth]{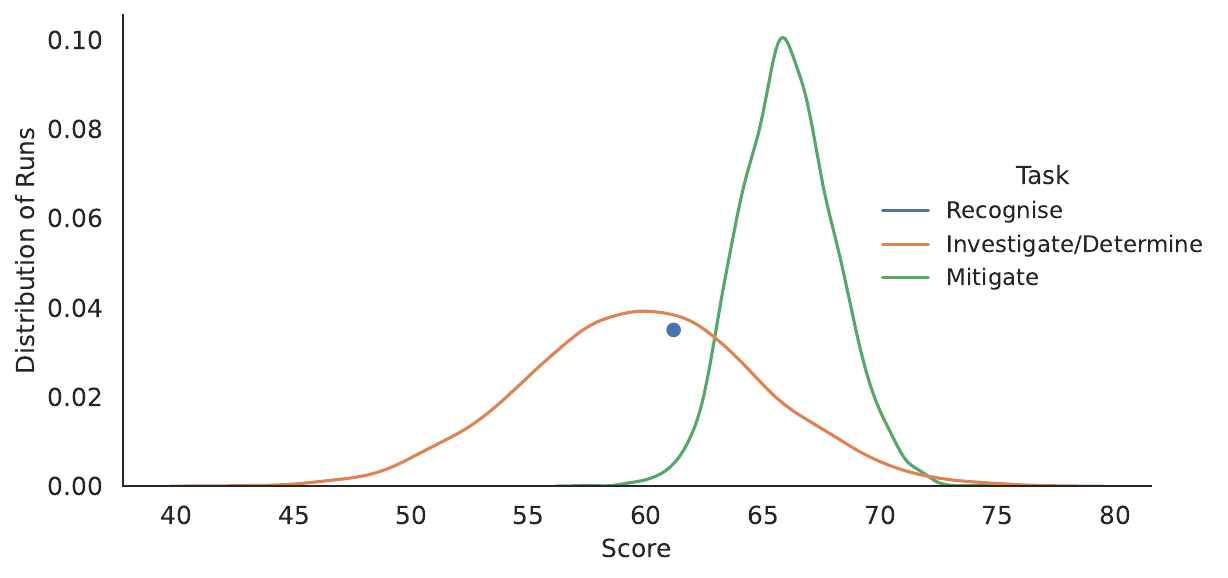}
        \caption{Various depths the effects of randomness can be addressed in the papers}
        \label{fig:distribution_of_tasks}
    \end{subfigure}
    \hfill
    \begin{subfigure}[b]{0.49\textwidth}
        \centering
        \includegraphics[width=\textwidth]{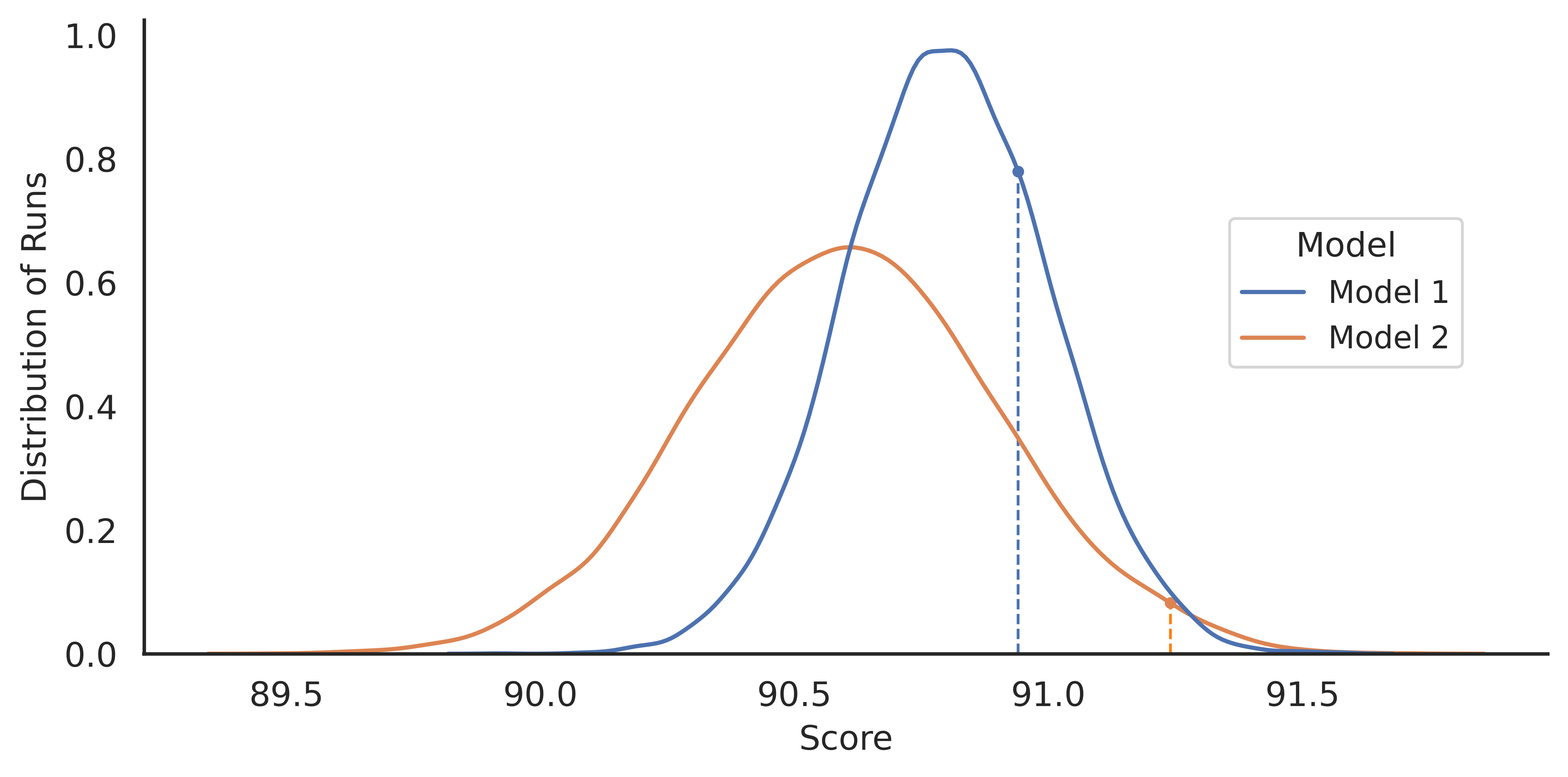}
        \caption{Comparison based on a single value and based on distribution of the results, inspired by \cite{reimers_reporting_2017}}
        \label{fig:distribution_comparison}
    \end{subfigure}
    \caption{(a) The effects of randomness can be addressed in various depths in the papers. (b) If not taken into consideration, randomness can introduce bias into comparison results, causing one approach to show better results only due to random chance and unintentional cherry-picking.}
    \label{fig:tasks_level_of_focus}
\end{figure}

If the uncontrolled randomness is not appropriately addressed in such low-data regimes, it can have significant, non-negligible negative consequences. In comparisons and benchmarks, changing only the random seed may lead to completely different model rankings~\cite{madaan2024quantifying, alzahrani2024benchmarks}. It may prohibit an objective performance comparison of newly designed methods to the state-of-the-art baselines, as it makes it difficult to conclude if a specific change made a meaningful difference or if the model just "got lucky" (especially when the difference between their performance is similar to the deviation caused by the randomness). In such cases, a method can be incorrectly denoted as state-of-the-art only based on a more favourable random chance~\cite{reimers_reporting_2017} (as illustrated in Figure~\ref{fig:distribution_comparison}). The uncontrolled randomness can unintentionally, but unfortunately also intentionally (by cherry-picking), create an imaginary perception of research progress. As a result, researchers even suggest that we should be very vigilant when relying on models and approaches that are significantly affected by sensitivity (e.g., large language models)~\cite{zong2024foolvisionandlanguage}. Finally, randomness has been identified as a significant obstacle that negatively affects reproducibility~\cite{albertoni2023reproducibility, chen2022towards}.

The effects of randomness are addressed in various depths - from a pure {\color{plt_blue}\textit{recognition}} of the effects, through a deeper {\color{plt_orange}\textit{investigation}} and {\color{plt_orange}\textit{determining}} the origin of the effects, up to their partial or full {\color{plt_green}\textit{mitigation}} (illustrated in the Figure~\ref{fig:tasks_level_of_focus}). First, the effects are simply recognised, resulting in the reporting of a more representative, albeit single, aggregated value from multiple training and evaluation runs (denoted as ``{\color{plt_blue}Recognise}''). As no further analysis is provided, this still leads to a biased comparison between approaches, as one approach may show better results only due to random chance and unintentional cherry-picking (illustrated in Figure~\ref{fig:distribution_comparison}). Another group of works perform analyses of the effects of randomness, estimate the distribution of results from multiple runs and compare the distributions (denoted as ``{\color{plt_orange}Investigate/Determine}''). The effects of the randomness are often analysed in more detail in order to determine the real origin of randomness in the training process (e.g., under-specification, which causes slightly different parameter initialisation to converge to different minima). As the distribution is only estimated, the deviation is still not reduced. The effects of randomness are fully addressed when they are mitigated, reducing the deviation in the distribution of results (denoted as ``{\color{plt_green}Mitigate}''). However, effective mitigation requires an understanding of the effects, their importance, and the origin of randomness provided by the analysis of the effects of randomness (provided by the ``{\color{plt_orange}Investigate/Determine}'' task).

In this paper, we conduct a comprehensive survey of 415 papers that address the effects of randomness, either by simply recognising the effects of randomness on stability (N=254), or by investigating/determining the characteristics of these effects in more detail and/or exploring the appropriate means of their mitigation (N=161)\footnote{The digital appendix with a full list of papers is available at \url{https://kinit.sk/public/acm-csur-sensitivity-survey.html}}. We provide a more in-depth analysis of the 161 papers that address the effects of randomness in more detail. Finally, we aggregate findings from this analysis and use them to identify seven challenges and open problems related to the study of the effects of randomness, while providing future directions on how to address them. For the purpose of this survey, we follow the PRISMA methodology~\cite{moher_2009_PRISMA, page_2021_prisma_updated} to systematically identify the papers in the defined scope.

This is the first survey of its kind that provides a comprehensive and systematic literature review specifically focused on the impact of the effects of randomness on stability across various approaches for dealing with limited labelled data. As opposed to previous papers that provide an overview of possible \textit{randomness factors} and investigate these effects \cite{pham_problems_2021, gundersen2022sources_of_irreproducibility, song2023comprehensive_survey_of_fsl}, provide an overview for a single task (e.g., mitigating the effects)~\cite{xu2024context}, or only recognise the problem randomness as an open problem~\cite{song2023comprehensive_survey_of_fsl, liu2021pre}, we focus on all tasks for addressing the effects of randomness, including investigation, but also determining their origin, mitigation, and benchmarking in the presence of these effects of randomness.

The purpose of this survey is to emphasise the importance of this research area, which so far has not received the appropriate level of attention from researchers and practitioners. First, it should serve existing or new researchers who are explicitly tackling the topic of randomness and its effects on the stability of \textit{learning with limited labelled data} to support their research (by providing an overview of the state-of-the-art, identification of open problems and future directions). Secondly, its purpose is to inform researchers and practitioners utilising the \textit{learning with limited labelled data} about the consequences of unaddressed randomness and how to effectively prevent them. To better achieve these purposes, we summarised all the identified papers along with their categorisation in the \href{https://kinit.sk/public/acm-csur-sensitivity-survey.html}{digital appendix of this survey} (which is further described in Section A of the supplementary material).

The rest of the paper is structured as follows. In Section~\ref{sec:background}, we describe the scope of the survey, the difference to existing surveys and the methodology used for identifying relevant papers. Taxonomy applied to analyse and categorise the identified papers is described in Section~\ref{sec:taxonomy}. Afterwards, we describe the different tasks for addressing the effects of randomness, categorised based on our taxonomy, in Sections~\ref{sec:investigation},~\ref{sec:det_source_of_randomness},~\ref{sec:mitigation} and~\ref{sec:benchmark-comparison}, with a summary of the findings at the end of each such section. The identified challenges and open problems are described in Section~\ref{sec:open-problems}. Finally, the survey is concluded in Section~\ref{sec:conclusion} with a summary of our contributions and findings.

\section{Background}
\label{sec:background}

\subsection{Scope of the Survey}
\label{sec:scope}

In this survey, we focus on a specific part of \textit{stability} and \textit{sensitivity} of machine learning approaches. We focus on papers that explore the effects of small perturbations to \textbf{multiple inputs} of the training process on the final performance after the training process is done. A similar, but at the same time quite distinct property is the \textit{robustness} of the model. It deals with a more large scale or systematic changes in the input, such as distribution shift or the presence of adversarial examples. As the \textit{adversarial robustness} is already extensively studied in the context of learning with limited labelled and the effects are often systematic, we consider it to be out of scope for this survey. Finally, we view the \textit{stability} to be a subset of the \textit{reproducibility} of the model. Besides randomness in the training process, reproducibility is impacted by other factors such as selective reporting, biases, data and source code availability, and many others as specified in~\cite{gundersen2022sources_of_irreproducibility}. These additional factors are out of scope for this survey.

Out of a wide spectrum of machine learning approaches, we specifically focus on \textit{learning with limited labelled data}, which is designed to train or prompt a model with only a limited number of labelled samples (typically 0-50 per class). Not requiring large sets of labelled data, these approaches are usable across a broader set of domains, making them also more popular. Due to this popularity, they also cover almost the whole research area of addressing the effects of randomness on stability, as there are almost no other works focusing on approaches that utilise information from unlabelled sets of data. Our focus is on five prevalent groups of approaches: 1) \textit{meta-learning}; 2) \textit{language model (LM) fine-tuning}; 3) \textit{prompting/in-context learning}; 4) \textit{prompt-based learning}; and 5) \textit{parameter-efficient fine-tuning (FT)}. Even though we do not explicitly focus on a specific modality, the popularity of the approaches impacts the scope of this survey. The most prominent modality is text, followed by images, as the majority of the papers focus on language model \textit{fine-tuning} and \textit{in-context learning}. Finally, even though we focus on the setting of limited data, we also include works that are relevant for this survey that utilise large sets of data. However, due to the search query used, such papers are not systematically covered and may be incomplete (even though we specifically designed one step in the methodology to catch these papers).

\subsection{Existing Surveys}

To the best of our knowledge, there are no existing surveys addressing the effects of randomness on stability across all tasks ranging from recognise to mitigate. However, there are relevant surveys on related topics that partially touch on the topics of \textit{stability} or \textit{sensitivity}. First of all, some papers provide an overview of possible sources of randomness, although with a focus on replicability~\cite{gundersen2022sources_of_irreproducibility}. In some surveys on \textit{learning with limited labelled data} and/or few-shot learning, the sensitivity to the effects of randomness is already mentioned as a problem that should be addressed~\cite{dong2022survey, song2023comprehensive_survey_of_fsl, liu2021pre, kaddour2023challenges}, but no additional details are provided. Finally, the most recent relevant survey~\cite{xu2024context} focuses on sample selection strategies for mitigating in-context learning sensitivity to sample choice. While relevant, it provides a significantly narrower perspective than the one given by this comprehensive survey. 

To fill this gap, we aim to conduct the first comprehensive overview of the effects of randomness on the stability of learning with limited labelled data and provide novel perspectives that are not easily visible from the standard analysis of related works.

\begin{figure}[tbh]
    \centering
    \includegraphics[width=0.90\textwidth]{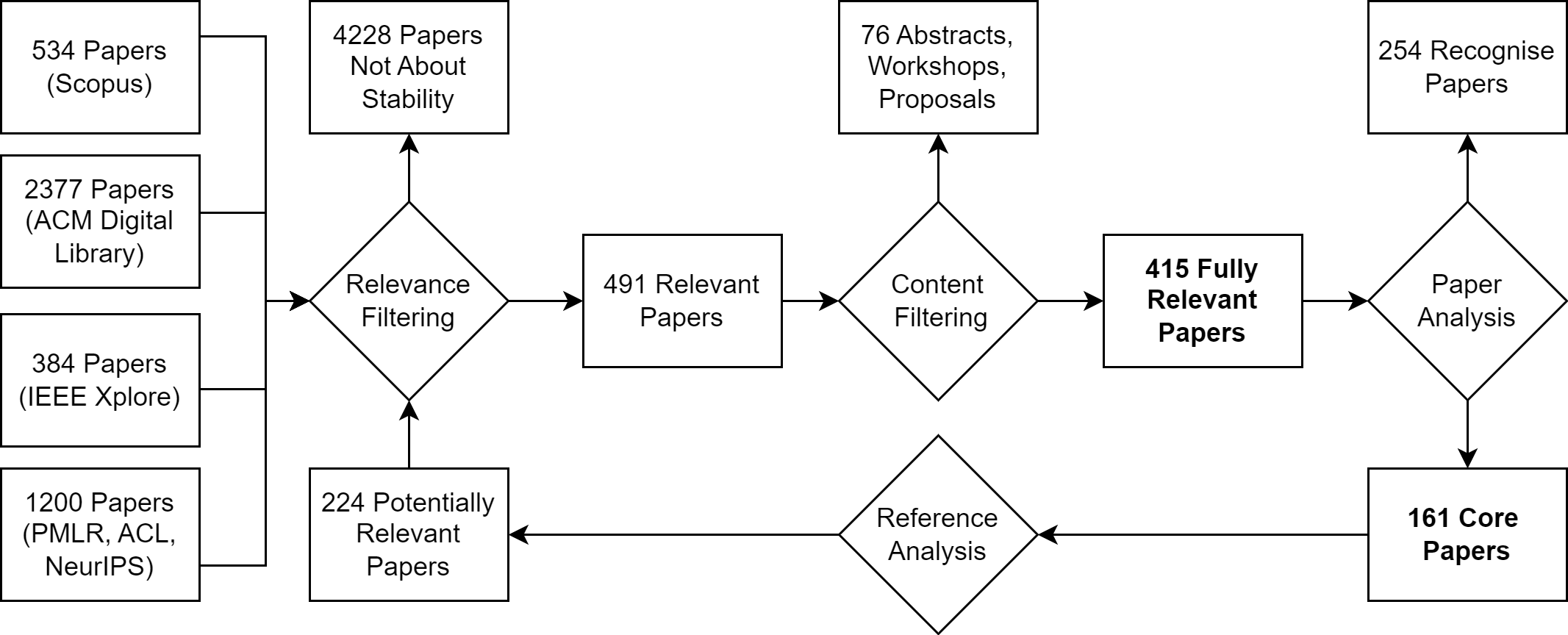}
    \caption{Process for identifying and categorising papers for this survey. Due to strong clustering effects, additional papers are identified using a reference analysis on the most relevant identified papers, i.e., including papers cited in them, as well as the ones that cite them.}
    \label{fig:paper_identification_process}
\end{figure}

\subsection{Survey Methodology}

To systematically identify the set of relevant papers for our comprehensive survey, we follow the PRISMA methodology~\cite{moher_2009_PRISMA, page_2021_prisma_updated}, performing the following steps (as illustrated in Figure \ref{fig:paper_identification_process}):
\begin{enumerate}
    \item \textbf{Identification of relevant papers.} A keyword search query is defined based on the related terms for our defined scope identified in different papers. Multiple digital libraries are searched using the defined query to identify a set of potentially relevant papers. 
    \item \textbf{Relevance filtering.} The potentially relevant papers are categorised into irrelevant and relevant based on the context the defined keywords are used in. 
    \item \textbf{Content filtering.} The non-full papers or papers released in unrelated venues, such as extended abstracts, proposals for talks or papers appearing in unrelated workshops not dealing with stability, are filtered out. 
    \item \textbf{Paper analysis.} The papers are analysed based on their depth of addressing the randomness effects. The papers that only recognise the problem of stability without further investigation are referred to as \textit{recognise} papers. The remaining papers are referred to as \textit{core} papers.
    \item \textbf{Reference analysis.} To discover additional potentially relevant papers (especially those relevant for the survey that do not necessarily deal with limited data), the papers cited in the \textit{core} papers; as well as papers citing the \textit{core} papers, are examined and their relevance is determined based on their title and abstract. The papers deemed potentially relevant are an additional input into the second step of this process.
    \item \textbf{Analysis and categorisation.} The identified \textit{core} papers are analysed and categorised based on the taxonomy (as defined in Section~\ref{sec:taxonomy}) to provide a comprehensive overview and to identify challenges and open problems.
\end{enumerate}

\begin{figure}[tbh]
    \centering
    \begin{subfigure}[b]{0.49\textwidth}
        \centering
        \includegraphics[width=\textwidth]{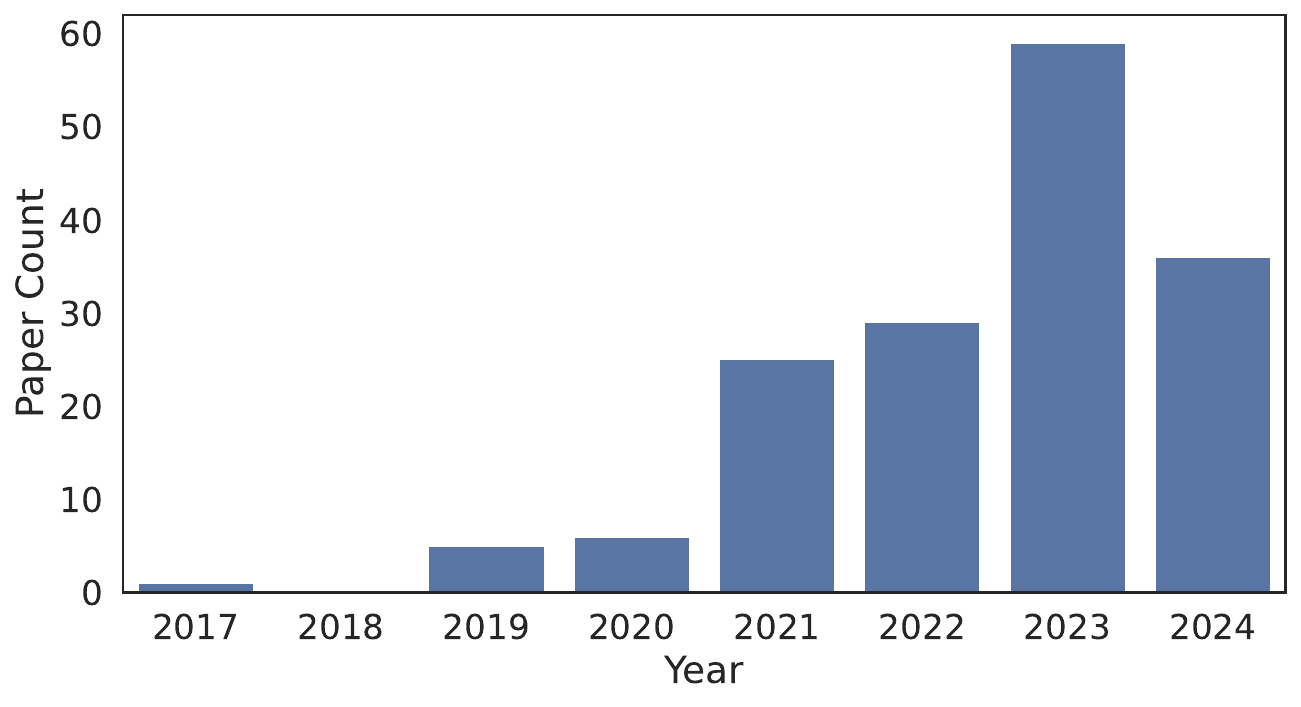}
        \caption{Core papers only}
        \label{fig:papers_through_years_core}
    \end{subfigure}
    \hfill
    \begin{subfigure}[b]{0.49\textwidth}
        \centering
        \includegraphics[width=\textwidth]{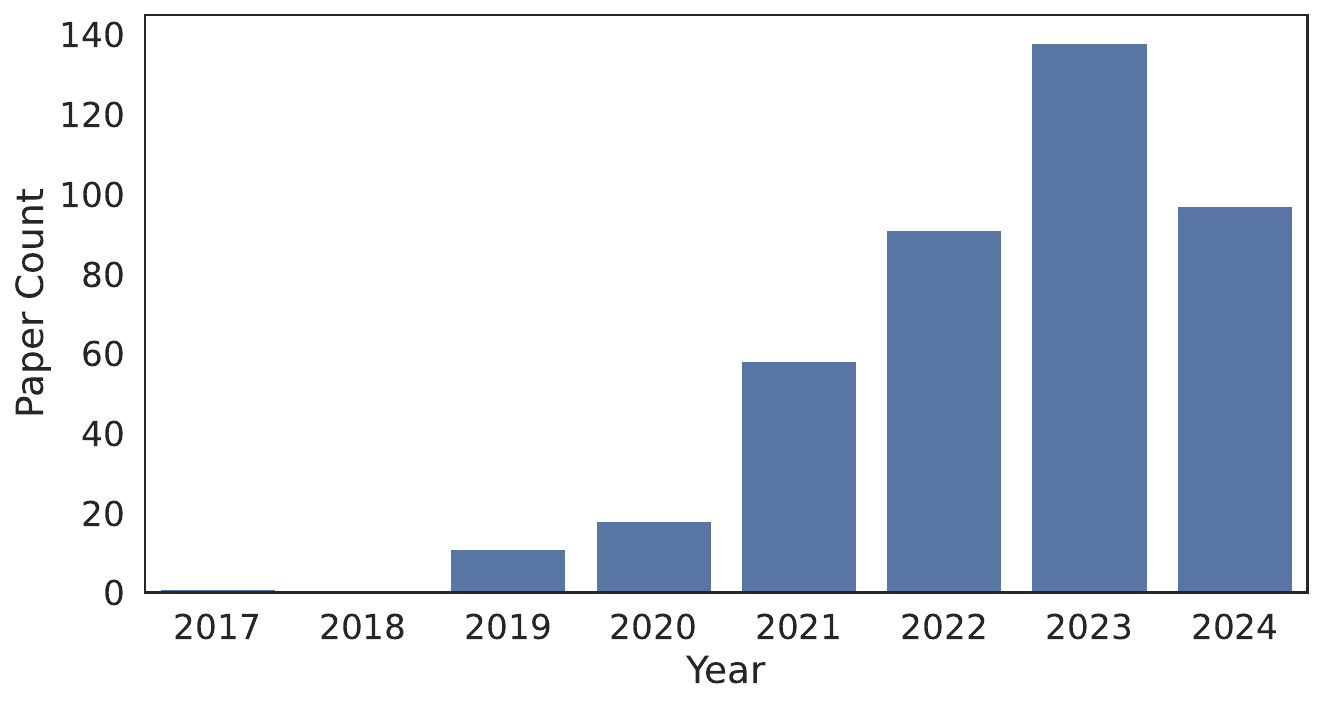}
        \caption{All papers}
        \label{fig:papers_through_years_all}
    \end{subfigure}
    \caption{Number of papers dealing with randomness grouped by the year. Figure \ref{fig:papers_through_years_core} shows only the \textit{core} papers that focus on addressing the effects of randomness in more detail, while Figure \ref{fig:papers_through_years_all} also includes the papers that only \textit{recognise} the problem. The problem started to attract attention only in recent years. Year 2024 covers papers published until July 2024.}
    \label{fig:papers_through_years}
\end{figure}

As a result, 415 papers are identified --- out of them, 254 papers are classified as \textit{recognise} papers, and 161 as \textit{core} papers. The number of identified papers grouped by the publication year is presented in Figure \ref{fig:papers_through_years}. The categorisation of the selected \textit{core} papers is showcased in Table \ref{tab:core_paper_categorisation}. Moreover, the full list of all identified papers (including \textit{recognise} papers), with their categorisation and additional metadata is available in the \href{https://kinit.sk/public/acm-csur-sensitivity-survey.html}{digital appendix}. For a more detailed implementation of the survey methodology, see the supplementary material (Section C).

\begin{table}[t]
\centering
\tiny
\caption{A showcase of the selected \textit{core} papers and their categorisation based on the proposed taxonomy. The full categorisation, along with the \textit{recognise} papers is available in the \href{https://kinit.sk/public/acm-csur-sensitivity-survey.html}{digital appendix}.}
\label{tab:core_paper_categorisation}
\renewcommand*{\arraystretch}{1.01}
%\resizebox{\linewidth}{!}{
\tabcolsep=0.12cm
\begin{tabular}{c|cccccc|ccccccccccccccc|cccc|cc|c|l}
\toprule
\multirow{2}{*}{\rotatebox[origin=l]{-90}{\textbf{References}}}
 & \multicolumn{6}{c|}{\textbf{ML Approach}} 
 & \multicolumn{15}{c|}{\textbf{Randomness Factor}} 
 & \multicolumn{4}{c|}{\textbf{Performed Tasks}} 
 & \multicolumn{2}{c|}{\textbf{Modality}} 
 & \multirow{2}{*}{\rotatebox[origin=l]{-90}{\textbf{Interactions}}} 
 & \multirow{2}{*}{\rotatebox[origin=l]{-90}{\textbf{Out-Of-Distribution}}} \\ 
 
 & \multicolumn{1}{l}{\rotatebox[origin=l]{-90}{Machine Learning}} 
 & \multicolumn{1}{l}{\rotatebox[origin=l]{-90}{Meta-Learning}} 
 & \multicolumn{1}{l}{\rotatebox[origin=l]{-90}{LM Fine-Tuning}} 
 & \multicolumn{1}{l}{\rotatebox[origin=l]{-90}{In-Context Learning}} 
 & \multicolumn{1}{l}{\rotatebox[origin=l]{-90}{Prompt-Based Learning}}
 & \multicolumn{1}{l|}{\rotatebox[origin=l]{-90}{Parameter-Efficient FT}}
 & \multicolumn{1}{l}{\rotatebox[origin=l]{-90}{Label Selection}} 
 & \multicolumn{1}{l}{\rotatebox[origin=l]{-90}{Data Split}} 
 & \multicolumn{1}{l}{\rotatebox[origin=l]{-90}{Task Choice}} 
 & \multicolumn{1}{l}{\rotatebox[origin=l]{-90}{Data Choice}} 
 & \multicolumn{1}{l}{\rotatebox[origin=l]{-90}{Order of Data}} 
 & \multicolumn{1}{l}{\rotatebox[origin=l]{-90}{Parameter Initialisation}} 
 & \multicolumn{1}{l}{\rotatebox[origin=l]{-90}{Random Seed}} 
 & \multicolumn{1}{l}{\rotatebox[origin=l]{-90}{Implementation}} 
 & \multicolumn{1}{l}{\rotatebox[origin=l]{-90}{Number of Samples}} 
 & \multicolumn{1}{l}{\rotatebox[origin=l]{-90}{Hyperparameters}} 
 & \multicolumn{1}{l}{\rotatebox[origin=l]{-90}{Augmentation}} 
 & \multicolumn{1}{l}{\rotatebox[origin=l]{-90}{Noise}}
 & \multicolumn{1}{l}{\rotatebox[origin=l]{-90}{Prompt}}
 & \multicolumn{1}{l}{\rotatebox[origin=l]{-90}{Model Randomness}}
 & \multicolumn{1}{l|}{\rotatebox[origin=l]{-90}{Approach Specific}} 
 & \multicolumn{1}{l}{\rotatebox[origin=l]{-90}{Investigate + Evaluate}}
 & \multicolumn{1}{l}{\rotatebox[origin=l]{-90}{Determine}} 
 & \multicolumn{1}{l}{\rotatebox[origin=l]{-90}{Mitigate}} 
 & \multicolumn{1}{l|}{\rotatebox[origin=l]{-90}{Benchmark + Compare + Report~~}} 
 & \multicolumn{1}{l}{\rotatebox[origin=l]{-90}{Text}} 
 & \multicolumn{1}{l|}{\rotatebox[origin=l]{-90}{Image}} 
 &  &  \\ \hline

\cite{bouthillier_accounting_2021} & $\sqrt{}$ &  &  &  & & & $\sqrt{}$ & $\sqrt{}$ &  &  & $\sqrt{}$ & $\sqrt{}$ &  &  &  & $\sqrt{}$ & $\sqrt{}$ & $\sqrt{}$ & & &  & $\sqrt{}$ &  &  & $\sqrt{}$ & $\sqrt{}$ &  & $\sqrt{}$ & $\times$  \\ \hline

\cite{agarwal_sensitivity_2021} &  & $\sqrt{}$ &  & & &  &  &  &  & $\sqrt{}$ &  &  &  &  &  &  &  &  & & &  & $\sqrt{}$ & $\sqrt{}$ & $\sqrt{}$ &  &  & $\sqrt{}$ & $\times$ & $\times$ \\ \hline
 
\cite{boquet_reproducibility_2019} &  & $\sqrt{}$ &  & & &  &  &  &  &  &  &  & $\sqrt{}$ & $\sqrt{}$ &  & $\sqrt{}$ &  &  & & &   & $\sqrt{}$ &  &  &  &  & $\sqrt{}$ & $\times$ & $\times$ \\ \hline
 
\cite{lu_fantastically_2022} &  &  &  & $\sqrt{}$ & &  &  &  &  &  & $\sqrt{}$ & - &  &  & $\sqrt{}$ &  &  &  & & &  & $\sqrt{}$ & $\sqrt{}$ & $\sqrt{}$ &  & $\sqrt{}$ &  & $\times$ & $\times$ \\ \hline
 
\cite{zhao_closer_2021} &  &  & $\sqrt{}$ & & &  &  &  &  & $\sqrt{}$ &  &  & $\sqrt{}$ &  & $\sqrt{}$ &  &  &  & & &  & $\sqrt{}$ & $\sqrt{}$ & $\sqrt{}$ &  & $\sqrt{}$ &  & $\sqrt{}$ & $\times$ \\ \hline
 
\cite{osahor_ortho-shot_2022} &  & $\sqrt{}$ &  & & &   &  &  & $\sqrt{}$ & $\sqrt{}$ &  &  &  &  &  &  &  &  &  &  & & &  & $\sqrt{}$ &  &  & $\sqrt{}$ & $\times$ & $\times$ \\ \hline
 
%\cite{zhang_review_2021} &  &  & $\sqrt{}$ & & &   &  &  &  &  & $\sqrt{}$ &  &  &  &  &  &  &  & & &  &  &  & $\sqrt{}$ &  & $\sqrt{}$ &  & $\times$ & $\times$ \\ \hline
 
\cite{min_noisy_2022} &  &  &  & $\sqrt{}$ & & $\sqrt{}$ &  &  &  &  & $\sqrt{}$ & - & $\sqrt{}$ &  & $\sqrt{}$ &  &  &  & & &  &  &  & $\sqrt{}$ &  & $\sqrt{}$ &  & $\times$ & $\times$ \\ \hline
 
%\cite{chen_meta-learning_2022} &  & $\sqrt{}$ &  & $\sqrt{}$ & $\sqrt{}$ &  &  &  &  & $\sqrt{}$ & $\sqrt{}$ &  &  &  &  &  &  &  & $\sqrt{}$ & & &  &  & $\sqrt{}$ &  & $\sqrt{}$ &  & $\times$ & $\times$ \\ \hline
 
%\cite{winata_language_2021} &  &  &  & $\sqrt{}$ & &  &  &  &  &  & $\sqrt{}$ & - &  &  &  &  &  &  &  & &  & $\sqrt{}$ &  &  &  & $\sqrt{}$ &  & $\times$ & $\times$ \\ \hline
 
\cite{chang_training_2021} &  &  & $\sqrt{}$ & & &   & $\sqrt{}$ &  &  &  &  &  &  &  & $\sqrt{}$ &  &  &  & & &   &  &  & $\sqrt{}$ &  & $\sqrt{}$ &  & $\times$ & $\times$ \\ \hline
 
\cite{schick_its_2021} &  &  &  & $\sqrt{}$ & $\sqrt{}$ &  &  &  &  & $\sqrt{}$ &  & - & $\sqrt{}$ &  &  &  &  & & &   &  & $\sqrt{}$ &  &  & $\sqrt{}$ & $\sqrt{}$ &  & $\times$ & $\times$ \\ \hline
 
\cite{jundi_how_2022} &  &  & $\sqrt{}$ & & &   & $\sqrt{}$ &  &  &  &  & $\sqrt{}$ &  &  & $\sqrt{}$ &  &  & & &   &  & $\sqrt{}$ &  & $\sqrt{}$ &  & $\sqrt{}$ &  & $\sqrt{}$ & $\times$ \\ \hline
 
\cite{wu_zero-shot_2022} &  &  & $\sqrt{}$ & & &   &  &  &  &  &  &  & $\sqrt{}$ &  &  &  &  &  & & &   &  & $\sqrt{}$ &  &  & $\sqrt{}$ &  & $\times$ & $\times$ \\ \hline
 
\cite{yang_efficient_2022} &  & $\sqrt{}$ & & &  &  &  &  & $\sqrt{}$ & $\sqrt{}$ &  &  &  &  &  &  &  &  &  & & &   &  & $\sqrt{}$ &  &  & $\sqrt{}$ & $\times$ & $\times$ \\ \hline
 
\cite{zhao_calibrate_2021} &  &  &  & $\sqrt{}$ & & &   &  &  & $\sqrt{}$ & $\sqrt{}$ & - &  &  & $\sqrt{}$ &  &  &  & $\sqrt{}$ & &  & $\sqrt{}$ & $\sqrt{}$ & $\sqrt{}$ &  & $\sqrt{}$ &  & $\sqrt{}$ & $\times$ \\ \hline
 
\cite{setlur_two_2021} &  & $\sqrt{}$ &  & & &   &  &  & $\sqrt{}$ &  &  &  &  &  &  &  &  &  &  &  & & &   &  & $\sqrt{}$ &  & $\sqrt{}$ & $\times$ & $\sqrt{}$ \\ \hline
 
\cite{dror_deep_2019} & $\sqrt{}$ &  & & &   &  &  &  &  &  & $\sqrt{}$ & $\sqrt{}$ & $\sqrt{}$ &  &  & $\sqrt{}$ &  &  & & &   &  &  &  & $\sqrt{}$ & $\sqrt{}$ &  & $\times$ & $\times$ \\ \hline
 
\cite{dodge_fine-tuning_2020} &  &  & $\sqrt{}$ & & &   &  &  &  &  & $\sqrt{}$ & $\sqrt{}$ & $\sqrt{}$ &  &  &  &  & & &  & $\sqrt{}$ & $\sqrt{}$ &  & $\sqrt{}$ &  & $\sqrt{}$ &  & $\sqrt{}$ & $\times$ \\ \hline
 
\cite{setlur_is_2021} &  & $\sqrt{}$ &  & & &   & $\sqrt{}$ &  &  & $\sqrt{}$ &  &  &  &  &  &  &  & & &   &  & $\sqrt{}$ &  &  &  &  & $\sqrt{}$ & $\times$ & $\times$ \\ \hline
 
\cite{mosbach_stability_2021} &  &  & $\sqrt{}$ & & &   &  &  &  &  &  &  & $\sqrt{}$ &  &  & $\sqrt{}$ & & &   &  &  &  & $\sqrt{}$ & $\sqrt{}$ &  & $\sqrt{}$ &  & $\times$ & $\times$ \\ \hline
 
\cite{kumar_reordering_2021} &  &  &  & $\sqrt{}$ & & &   &  &  &  & $\sqrt{}$ & - &  &  &  &  &  &  &  & & &   &  & $\sqrt{}$ &  & $\sqrt{}$ &  & $\times$ & $\times$ \\ \hline
 
\cite{reimers_reporting_2017} & $\sqrt{}$ &  & & &   &  &  &  &  &  &  &  & $\sqrt{}$ &  &  & $\sqrt{}$ &  &  & &  &  & $\sqrt{}$ &  &  & $\sqrt{}$ & $\sqrt{}$ &  & $\times$ & $\times$ \\ \hline
 
\cite{zhang_revisiting_2021} &  &  & $\sqrt{}$ & & &   &  &  &  &  &  & $\sqrt{}$ &  &  &  & $\sqrt{}$ &  & & &  & $\sqrt{}$ &  & $\sqrt{}$ & $\sqrt{}$ &  & $\sqrt{}$ &  & $\sqrt{}$ & $\times$ \\ \hline
 
\cite{liu_what_2022} &  &  &  & $\sqrt{}$ & &  & $\sqrt{}$ &  &  & $\sqrt{}$ & $\sqrt{}$ & - &  &  & $\sqrt{}$ &  & & &   &  &  & $\sqrt{}$ & $\sqrt{}$ & $\sqrt{}$ &  & $\sqrt{}$ &  & $\times$ & $\times$ \\ \hline
 
\cite{sellam_multiberts_2022} &  &  & $\sqrt{}$ & & &   &  &  &  & $\sqrt{}$ & $\sqrt{}$ & $\sqrt{}$ & $\sqrt{}$ &  &  & & &   &  &  &  & $\sqrt{}$ &  &  & $\sqrt{}$ & $\sqrt{}$ &  & $\sqrt{}$ & $\times$ \\ \hline
 
\cite{mccoy_berts_2020} &  &  & $\sqrt{}$ & & &   &  &  &  &  & $\sqrt{}$ & $\sqrt{}$ &  &  &  &  &  & & &   &  & $\sqrt{}$ & $\sqrt{}$ & $\sqrt{}$ &  & $\sqrt{}$ &  & $\times$ & $\sqrt{}$ \\ \hline
 
\cite{zhou_curse_2020} &  &  & $\sqrt{}$ & & &   &  &  &  &  &  &  & $\sqrt{}$ &  &  &  & & &   &  &  & $\sqrt{}$ & $\sqrt{}$ & $\sqrt{}$ &  & $\sqrt{}$ &  & $\times$ & $\sqrt{}$ \\ \hline
 
\cite{bouthillier_unreproducible_2019} & $\sqrt{}$ &  & & &   &  &  &  &  &  & $\sqrt{}$ & $\sqrt{}$ &  &  &  &  & & &   &  &  & $\sqrt{}$ &  &  & $\sqrt{}$ &  & $\sqrt{}$ & $\times$ & $\times$ \\ \hline
 
\cite{dehghani_benchmark_2021} & $\sqrt{}$ & & &   &  &  &  & $\sqrt{}$ & $\sqrt{}$ & $\sqrt{}$ &  &  & $\sqrt{}$ &  & & &  &  &  &  &  & $\sqrt{}$ &  &  & $\sqrt{}$ & $\sqrt{}$ & $\sqrt{}$ & $\times$ & $\times$ \\ \hline
 
\cite{zhong_are_2021} &  &  & $\sqrt{}$ & & &   &  &  &  &  & $\sqrt{}$ & $\sqrt{}$ & $\sqrt{}$ &  &  &  &  & & &   &  & $\sqrt{}$ &  &  & $\sqrt{}$ & $\sqrt{}$ &  & $\sqrt{}$ & $\times$ \\ \hline
 
\cite{dauphin_metainit_2019} & $\sqrt{}$ & & &   &  &  &  &  &  &  &  & $\sqrt{}$ &  &  &  &  &  & & &   &  &  &  & $\sqrt{}$ &  &  & $\sqrt{}$ & $\times$ & $\times$ \\ \hline
 
\cite{damour_underspecification_2020} & $\sqrt{}$ &  & $\sqrt{}$ & & &   &  &  &  &  &  &  & $\sqrt{}$ &  &  &  & & &   &  &  &  & $\sqrt{}$ &  &  & $\sqrt{}$ & $\sqrt{}$ & $\times$ & $\times$ \\ \hline
 
\cite{khurana_how_2021} &  &  & $\sqrt{}$ & & &   &  &  &  &  &  &  & $\sqrt{}$ &  &  &  &  &  & & &   &  &  & $\sqrt{}$ &  & $\sqrt{}$ &  & $\times$ & $\times$ \\ \hline
 
\cite{hua_noise_2021} &  &  & $\sqrt{}$ &  & &  &  &  &  &  & $\sqrt{}$ & $\sqrt{}$ & $\sqrt{}$ &  &  & & &   &  &  &  &  & $\sqrt{}$ & $\sqrt{}$ &  & $\sqrt{}$ &  & $\times$ & $\times$ \\ \hline
 
\cite{hidey_reducing_2022} &  &  & $\sqrt{}$ & & &  &  &  &  &  & $\sqrt{}$ & $\sqrt{}$ &  &  &  &  &  & & &   &  &  &  & $\sqrt{}$ &  & $\sqrt{}$ &  & $\times$ & $\times$ \\ \hline
 
\cite{pham_problems_2021} & $\sqrt{}$ &  & & &  &  &  &  &  &  & $\sqrt{}$ & $\sqrt{}$ &  & $\sqrt{}$ &  & & &   &  &  &  & $\sqrt{}$ &  &  &  &  & $\sqrt{}$ & $\times$ & $\times$ \\ \hline
 
\cite{cioba_how_2022} &  & $\sqrt{}$ & & &   &  & $\sqrt{}$ &  &  &  &  &  &  &  &  &  &  &  &  & $\sqrt{}$ &  & & &   &  &  & $\sqrt{}$ & $\times$ & $\sqrt{}$ \\ \hline
 
\cite{phang_sentence_2019} &  &  & $\sqrt{}$ & & &  &  &  &  &  &  &  & $\sqrt{}$ &  &  &  &  &  &  & & &   &  & $\sqrt{}$ &  & $\sqrt{}$ &  & $\times$ & $\sqrt{}$ \\ \hline
 
\cite{wang_variance-reduced_2021} &  & $\sqrt{}$ &  & & &   &  &  & $\sqrt{}$ & $\sqrt{}$ &  &  &  &  &  &  &  &  & & &  &  &  & $\sqrt{}$ &  & $\sqrt{}$ &  & $\times$ & $\times$ \\ \hline
 
\cite{zhang_fine-tuning_2022} &  &  & $\sqrt{}$ & & &   &  &  &  &  &  &  & $\sqrt{}$ &  &  &  &  &  & & &   &  &  & $\sqrt{}$ &  & $\sqrt{}$ &  & $\times$ & $\sqrt{}$ \\ \hline
 
%\cite{lee_mixout_2020} &  &  & $\sqrt{}$ & & &   &  &  &  &  &  &  & $\sqrt{}$ &  &  &  &  &  &  & &  &  &  & $\sqrt{}$ &  & $\sqrt{}$ &  & $\times$ & $\times$ \\ \hline
 
\cite{xu_raise_2021} &  &  & $\sqrt{}$ & & &   &  &  &  &  &  &  & $\sqrt{}$ &  &  &  &  &  & & &   &  &  & $\sqrt{}$ &  & $\sqrt{}$ &  & $\times$ & $\times$ \\ \hline
 
\cite{meng_tuning_2022} &  &  & $\sqrt{}$ & $\sqrt{}$ & & &   &  &  &  &  &  & $\sqrt{}$ &  & & &   &  &  & $\sqrt{}$ &  &  &  & $\sqrt{}$ &  & $\sqrt{}$ &  & $\times$ & $\times$ \\ \hline
 
\cite{zhang_active_2022} &  &  &  & $\sqrt{}$ & & &   &  &  & $\sqrt{}$ & $\sqrt{}$ & - &  &  &  &  & &  &  &  &  & $\sqrt{}$ & $\sqrt{}$ & $\sqrt{}$ &  & $\sqrt{}$ &  & $\sqrt{}$ & $\times$ \\ \hline
 
\cite{zheng_fewnlu_2022} &  &  & $\sqrt{}$ & & &   &  & $\sqrt{}$ &  &  &  &  &  &  &  & $\sqrt{}$ &  & & &   &  &  &  & $\sqrt{}$ & $\sqrt{}$ & $\sqrt{}$ &  & $\times$ & $\times$ \\ \hline
 
\cite{ye_crossfit_2021} &  & $\sqrt{}$ &  & & &   &  &  & $\sqrt{}$ &  &  &  & $\sqrt{}$ &  &  &  &  & & &   &  & $\sqrt{}$ &  &  & $\sqrt{}$ & $\sqrt{}$ &  &$\times$  & $\sqrt{}$ \\ \hline
 
\cite{bragg_flex_2021} &  &  & $\sqrt{}$ & $\sqrt{}$ & & &   &  &  &  &  &  & $\sqrt{}$ &  & $\sqrt{}$ & $\sqrt{}$ &  &  & & &   &  &  &  & $\sqrt{}$ & $\sqrt{}$ &  & $\times$ & $\times$ \\ \hline
 
\cite{mukherjee_clues_2021} &  &  & $\sqrt{}$ & $\sqrt{}$ & & &   & $\sqrt{}$ &  &  &  &  & $\sqrt{}$ &  & $\sqrt{}$ &  &  &  & & &   &  &  & $\sqrt{}$ & $\sqrt{}$ & $\sqrt{}$ &  & $\times$ & $\times$ \\ \hline
 
\cite{efrat_lmentry_2022} &  &  &  & $\sqrt{}$ & & &   &  &  & $\sqrt{}$ & $\sqrt{}$ & - &  &  &  &  &  &  & $\sqrt{}$ & &  & $\sqrt{}$ &  &  & $\sqrt{}$ & $\sqrt{}$ &  & $\times$ & $\times$ \\ \bottomrule

\end{tabular}%}

\end{table}

\section{Taxonomy for Literature Analysis and Categorisation}
\label{sec:taxonomy}

The identified \textit{core} papers are categorised based on multiple dimensions that describe their characteristics of interest. We define the following three separate primary properties that are used to categorise the papers: 1) what \textit{tasks are performed to address the randomness}; 2) \textit{randomness factors} addressed by the papers; and 3) what \textit{machine learning (ML) approach} the papers use. The \textit{performed tasks} property is the main property based on which we organise the rest of the survey, while the \textit{randomness factors} and \textit{machine learning approach} properties are mostly used to define the scope in this survey (see Section~\ref{sec:scope}). We provide the paper distribution across individual \textit{randomness factors} and specific \textit{machine learning approaches} for all \textit{core} papers in Table \ref{tab:factor-approach-mapping} (the paper distributions for specific tasks are included in the \href{https://kinit.sk/public/acm-csur-sensitivity-survey.html}{digital appendix}.

\begin{table}[]
\caption{Relation between the \textit{randomness factors} and \textit{machine learning approaches} across all analysed papers. We can observe different popularity of the \textit{randomness factors} across different \textit{machine learning approaches}. The majority focus is denoted in \textbf{bold}. For example, \textit{fine-tuning} mostly focuses on the random seeds.}
\label{tab:factor-approach-mapping}
\small
\resizebox{\linewidth}{!}{
\tabcolsep=0.15cm
\begin{tabular}{lcccccc}
\toprule
                     & All        & Meta-Learning    & Fine-Tuning & In-Context Learning & Prompt-Based Learning & PEFT \\
                     & (N=13)     & (N=14)           & (N=35)      & (N=96)             & (N=17)                & (N=10)         \\ \midrule
Label Selection      & 1          & 3                & 4           & 3                   & 1                      &               \\
Data Split           & 2          & 1                & 4           & 3                   & 1                      &               \\
Task Choice          & 1          & \textbf{7}       & N/A         &                     &                       &               \\
Data Choice          & 1          & \textbf{9}       & 6           & \textbf{51}         &\textbf{8}             &               \\
Random Seed          & \textbf{7} & 3                & \textbf{20} & 7                   & 3                     & 2              \\
Model Initialisation & \textbf{9} & 1                & 13          & N/A                & 2                     & 4              \\
Order of Data        & \textbf{8} & 2                & 11          & \textbf{37}         & 6                     & 2              \\
Implementation       & 5          & 1                &             &                     &                       &               \\
Number of Samples    &            & 3                & 9           & 17                  & 2                     & 2              \\
Hyperparameters      & 3          & 1                & 4           & 1                   &                       &               \\
Augmentation         & 3          &                  & 1           &                     &                       & 1              \\
Noise                & 1          &                  & 2           & 1                   &                       & 1              \\
Prompt               & N/A        & N/A              & N/A         & \textbf{46}         & \textbf{11}           & \textbf{5}    \\
Model Randomness     & 3          & 1                & 3           & 3                   &  2                    &  1             \\
Approach Specific    & N/A        & 1                & 5           & 3                   &                       &               \\
\bottomrule
\end{tabular}}
\end{table}

\subsection{Tasks Performed to Address the Randomness}

\begin{figure}[tbh]
    \centering
    \includegraphics[width=0.95\textwidth]{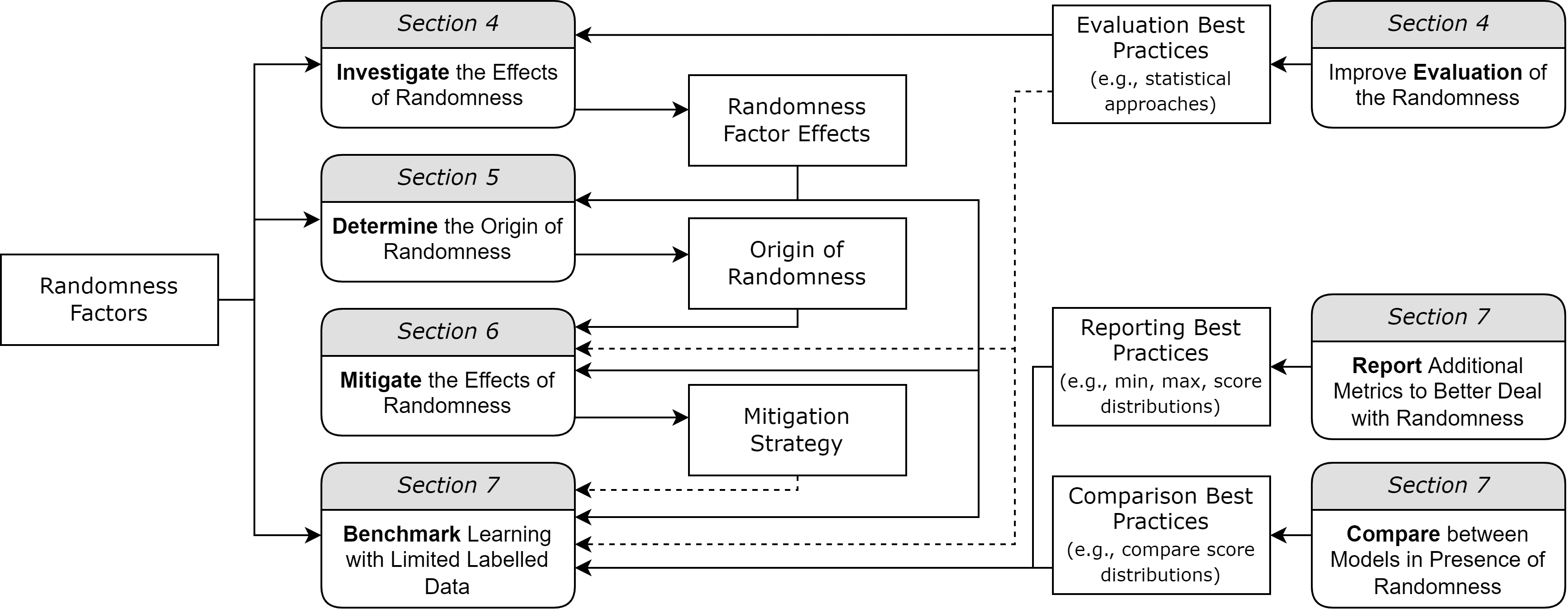}
    \caption{Different tasks for addressing the effects of randomness, along with their inputs, outputs and the relations between them. The dashed lines represent relations between the tasks that currently do not exist but need to be considered in the optimal state of addressing the effects of randomness, e.g., using more sophisticated evaluation when determining the effectiveness of mitigation strategies. The related tasks are grouped, e.g., \textit{evaluate} task near the \textit{investigate} task.}
    \label{fig:tasks-addressing-randomness}
\end{figure}

This property specifies the different tasks for addressing the effects of randomness that are performed in the papers (the relations between these tasks, along with their input and output are illustrated in Figure \ref{fig:tasks-addressing-randomness}). For this property, we define the following 7 values:
\begin{enumerate}
    \item \textit{Investigate} - the effects of randomness from different \textit{randomness factors} are investigated using specifically designed experiments. For example, the training and evaluation is done multiple times to determine the distribution of performance across multiple runs.
    \item \textit{Determine} - an effort to discover the origin of the randomness (different from the \textit{randomness factors}) is made by analysing the behaviour of the specific machine learning approach or model in the presence of randomness. For example, to determine why changing the order of samples causes a significant change in performance, the experiments that observe specific problems (e.g., label bias) are designed and analysed.
    \item \textit{Mitigate} - mitigation strategies for reducing the effects of randomness are proposed, used and their effectiveness is determined. For example, it is proposed to use more deterministic data sampling based on heuristics instead of random sampling.
    \item \textit{Benchmark} - a benchmark that is specifically designed to take into consideration the effects of randomness is proposed. For example, a benchmark that performs multiple train and test splits and evaluates on all of them.
    \item \textit{Evaluate} - a more sophisticated, usually statistical, framework for evaluating the experiments investigating the effects of randomness, or determining the effectiveness of mitigation strategies, is proposed. For example, instead of calculating a simple average value, it is proposed to estimate the distribution of results and evaluate the difference using statistical tests.
    \item \textit{Report} - it is proposed to report more than a single value when running and evaluating experiments and comparing approaches to better deal with the effects of randomness. Not necessarily dependent on experiments for addressing the effects of randomness, but also recommendations on how to improve the reporting of results overall when the randomness has a strong effect on performance. For example, instead of reporting a single average value, the minimum and maximum values over multiple runs should be reported as well.
    \item \textit{Compare} - better comparison strategies between different approaches in the presence of significant effects of randomness are proposed. For example, instead of comparing single average values, it is proposed to compare distributions of results using statistical tests.
    
\end{enumerate}

We aggregate the closely related tasks together and organise the survey based on these aggregated groups. Namely, we aggregate together the \textit{investigate} and \textit{evaluate} tasks, as the evaluation strategies are designed specifically for the investigation experiments. In addition, we aggregate the \textit{benchmark}, \textit{compare} and \textit{report} values, as they are closely related.

As the different papers can focus on different, often disjointed, tasks for addressing the effects of randomness, this property does not necessarily provide a distinct division of the papers. For example, the most common combination is to perform \textit{investigation}, \textit{determination} and \textit{mitigation} of a specific \textit{randomness factor} in a single paper. In such cases, the paper is mentioned in this survey multiple times, each time focusing only on the parts relevant to the specific task.

\subsection{Randomness Factors}
\label{sec:randomness_factors}

The \textit{randomness factors} property specifies what \textit{randomness factors} are addressed in the paper in some way (i.e., their effects are investigated or mitigated), but not \textit{factors} that are just mentioned (e.g., mention that other papers found the factor to be unimportant). Papers can focus on multiple \textit{factors} at the same time.

Multiple \textit{randomness factors} have already been identified and described in the existing works (e.g., \cite{pham_problems_2021, gundersen2022sources_of_irreproducibility}). Since the used terminology differs across the works, for the purpose of this survey, we established the grounds (in terms of common terminology and understanding) by collecting these \textit{randomness factors} and unifying their denomination and definition.

The identified \textit{randomness factors} can be categorised into three distinct groups: 1) \textit{factors} related to the input data; 2) \textit{factors} related to parameters of the model and the overall training process; 3) \textit{factors} related to the underlying implementation of deep learning frameworks and hardware; and 4) \textit{factors} that provide more systematic changes. 

As the \textit{randomness factors} related to the input data, we identify the following:
\begin{enumerate}
    \item \textit{Label selection} - specifies what data is considered to be annotated from a larger set of unlabelled data. Represents the real-world scenario with a limited labelled budget, where only specific samples are annotated. Has a potentially significant impact on the underlying data distribution.
    \item \textit{Data split} - randomness introduced by different splits of data into training, validation and testing set. Mainly impacts the information provided to the model during training.
    \item \textit{Task choice} - randomness introduced by choosing different tasks for training. This factor is mainly relevant for approaches that perform \textit{few-shot learning}, such as \textit{meta-learning}, or when using a benchmark dataset composed of different tasks.
    \item \textit{Data choice} - randomness introduced by the sampling of data the model uses for adaptation (\textit{meta-learning}), or as examples in prompts (\textit{in-context learning}).
    \item \textit{Order of data} - randomness introduced by the random shuffling of data in different training iterations. Mainly relevant for the \textit{in-context learning} (order of labelled samples in the prompt) and \textit{fine-tuning}. For \textit{meta-learning}, this factor represents the order in which the different tasks are presented during training.
    \item \textit{Noise} - randomness introduced by adding noise to the different samples. 
    \item \textit{Prompt format} - randomness introduced by small changes in the prompt formats, such as replacing words with synonyms or words without semantic meaning.
\end{enumerate}

Similarly, we identify the following \textit{randomness factors} corresponding to the randomness related to model parameters and the overall training process:
\begin{enumerate}
    \item \textit{Parameter initialisation} - randomness introduced by randomly initialising the parameters of the model, such as weights in neural networks. Mainly relevant for \textit{meta-learning} and \textit{language model fine-tuning}, as no initialisation is performed in \textit{in-context learning}. 
    \item \textit{Random seed} - randomness introduced by non-deterministic actions in the training process. Often used as an aggregated \textit{factor} for the parameter initialisation and the order of data \textit{randomness factors}.
    \item \textit{Model randomness} - non-deterministic operations in the model, such as the use of non-deterministic layers or regularisation.
    \item \textit{Approach specific parameters} - aggregates the randomness introduced by \textit{factors} specific for only some of the approaches (e.g., sampling strategy or number of iterations). 
\end{enumerate}

Randomness introduced by \textit{implementation of deep learning frameworks and hardware} is a special low-level group, which includes \textit{factors} such as scheduling and floating point precision, auto-selection of primitive operations, how parallel processes are used, changes introduced by different implementation of frameworks such as TensorFlow or PyTorch and their different software versions, but also many other \textit{factors} influencing reproducibility more than the stability of results \cite{pham_problems_2021, gundersen2022sources_of_irreproducibility}. Since these \textit{randomness factors} cannot be easily addressed, for this survey, we do not specifically distinguish between them and consider them jointly as one group.

Finally, we also consider \textit{factors} that provide more systematic changes and can be viewed as an edge case with regards to randomness. This group includes factors such as the randomness introduces by augmentation, hyperparameters or the number of samples. Besides these \textit{randomness factors}, we also consider \textit{systematic choices} as they can affect the sensitivity to the effects of randomness as well. These are often a result of the experimental setup and include choices such as the datasets and their characteristics (e.g., monolingual vs. multilingual), or the model size

Each \textit{randomness factor} is associated with a set of its \textit{randomness factor configurations} (sometimes referenced in short as \textit{configurations}). A single \textit{configuration} specifies one state from the set of all possible states the \textit{randomness factor} can attain. For example, a single \textit{randomness factor configuration} specifies one possible permutation of the samples (order), a single set of weights used for initialising the whole model, or a single random value to seed the random number generator. Using this definition, the set of \textit{randomness factor configuration} can be infinite for some \textit{factors} (e.g., initialisation weights) or finite, but extremely large (e.g., permutations of order).

The different \textit{randomness factors} may depend on each other, leading to \textit{interactions} between them~\cite{pecher2024sensitivity, weber-etal-2023-mind, chen-etal-2022-revisiting}. An example of such \textit{interactions} is the relation between data choice and order of data \textit{randomness factors}, where the choice of what data we use in training explicitly affects the order of samples. Therefore, such interactions can lead to negative, often unforeseen consequences, such as prohibiting reliable freezing of \textit{randomness factor configurations} (e.g., it is not possible to have a fixed order of samples when using different choices of data).

\subsection{Machine Learning Approach}
\label{sec:learning_with_limited_labelled_data}

Similarly, the \textit{machine learning approach} property specifies which group of approaches is addressed in the paper. Besides the machine learning approaches that define the scope of this survey (Section~\ref{sec:scope}), we also consider a generic \textit{machine learning}. This is a special value that contains papers that still deal with the effects of randomness, but only as part of overall supervised learning and do not focus on any of the specific machine learning approaches. The main idea, high-level overview and reference to popular approaches and comprehensive surveys of the machine learning approaches we work in this survey are included in the supplementary material (Section B).

\section{Investigating the problem of stability}
\label{sec:investigation}

In this task, the effects of randomness on the stability are investigated, their impact is estimated and the significance of this impact is determined. Each paper investigating the effects of randomness can be characterised by three main aspects: 1) \textit{scope} of the investigation; 2) investigation \textit{methodology}; and 3) \textit{results} from the investigation. The \textit{scope} of the investigation includes decisions about what \textit{factors} to investigate or across what models, datasets and modalities the investigation is performed. It also includes more nuanced specifics like what problem statement is considered (simple classification/regression or sequence generation), or from what setting the data comes from (in-distribution vs. out-of-distribution). Although these decisions have minimal impact on the \textit{methodology}, they strongly govern the obtain \textit{results} and their generality.

The investigation \textit{methodology} determines how the investigation is done. It provides answers to different problems and specifics of the investigation, such as how to perform the investigation (multiple fully randomised runs or keeping some parts static), how many investigation runs to use, how to address interactions between \textit{randomness factors} or how to evaluate the different runs. Again, each decision has a significant impact on the \textit{results} of the investigation, mainly how trustworthy and objective they are.

Finally, the \textit{results} of the investigation specify findings and the behaviour that was observed. Interpreting these \textit{results} is important for informing the further tasks for addressing the effects of randomness (i.e., determining the real origin of randomness and mitigating the effects). They can also be used to provide recommendations on the best use of models in practice in low-data regimes.

\subsection{Scope of the investigation}

\textbf{The majority of the papers focus on investigating how the randomness affects the language models} (41 out of 56) when they are used to deal with limited labelled data. Most of the focus is on \textit{in-context learning}~\cite{lu_fantastically_2022, schick_its_2021, zhao_calibrate_2021, liu_what_2022, zhang_active_2022, efrat_lmentry_2022} (32) and only a smaller focus is on \textit{fine-tuning of language models}~\cite{sellam_multiberts_2022, zhao_closer_2021, jundi_how_2022, dodge_fine-tuning_2020, mccoy_berts_2020, zhou_curse_2020, zhong_are_2021} (9) or \textit{prompt-based learning} (4). Parameter-efficient fine-tuning methods are not investigated at all (only recognised in~\cite{chen-etal-2022-revisiting}). The focus on investigating overall \textit{machine learning} approaches (mainly their use with few data and evaluation) \cite{bouthillier_accounting_2021, reimers_reporting_2017, bouthillier_unreproducible_2019, dehghani_benchmark_2021, pham_problems_2021} and the use of \textit{meta-learning} \cite{boquet_reproducibility_2019, agarwal_sensitivity_2021, setlur_is_2021, cioba_how_2022, ye_crossfit_2021, pecher2024sensitivity} is significantly lower (10 and 6 respectively). Only a fraction of papers consider multiple approaches at the same time~\cite{pecher2024sensitivity, webson-pavlick-2022-prompt, xu-etal-2023-multiinstruct, schick_its_2021}.

The significant focus on language models also translates to what modality is addressed the most. \textbf{The investigation of text data in NLP problems receives the most focus} (46 out of 56) \cite{bouthillier_accounting_2021, sellam_multiberts_2022, lu_fantastically_2022, zhao_closer_2021, schick_its_2021, jundi_how_2022, zhao_calibrate_2021, dodge_fine-tuning_2020, reimers_reporting_2017, liu_what_2022, mccoy_berts_2020, zhou_curse_2020, zhong_are_2021, zhang_active_2022, ye_crossfit_2021, efrat_lmentry_2022}, with images being mostly addressed only in \textit{meta-learning} and \textit{machine learning} approaches (but not all of them) \cite{boquet_reproducibility_2019, agarwal_sensitivity_2021, setlur_is_2021, bouthillier_unreproducible_2019, pham_problems_2021, cioba_how_2022}, with only 3 papers considering investigation across both image and text data \cite{dehghani_benchmark_2021, summers_nondeterminism_2021, zong2024foolvisionandlanguage}.

\textbf{The popularity and importance of \textit{randomness factors} heavily depend on the approach.} The most focus is dedicated to the data order \textit{randomness factor}, being the most popular for machine learning, fine-tuning and in-context learning~\cite{lu_fantastically_2022, zhao_calibrate_2021, zhang_active_2022, efrat_lmentry_2022, pezeshkpour2023large, zong2024foolvisionandlanguage}. The remaining \textit{randomness factors} are investigated only for specific approaches. For in-context and prompt-based learning, the most popular factors are sample choice and prompt format~\cite{sclar_quantifying_2023, weber-etal-2023-mind, reif-schwartz-2024-beyond, zhang2024impact} and, to a certain extent, the number of samples. Parameter initialisation and random seed are most popular for general machine learning and fine-tuning approaches~\cite{boquet_reproducibility_2019, zhao_closer_2021, schick_its_2021, reimers_reporting_2017, zhou_curse_2020, dehghani_benchmark_2021, ye_crossfit_2021, jundi_how_2022, liu_what_2022, chen-etal-2022-revisiting}. The implementation randomness factors are popular only for the general machine learning papers~\cite{boquet_reproducibility_2019, pham_problems_2021, summers_nondeterminism_2021, zhuang_randomness_2022, gundersen_reporting_2023}. The remaining randomness factors, such as the effects of augmentation~\cite{bouthillier_accounting_2021, summers_nondeterminism_2021}, choice of what data is labelled~\cite{bouthillier_accounting_2021, jundi_how_2022, setlur_is_2021, liu_what_2022, cioba_how_2022, pecher2024sensitivity}, split of data~\cite{bouthillier_accounting_2021, dehghani_benchmark_2021, pecher2024sensitivity}, noise~\cite{bouthillier_accounting_2021}, task choice for meta-learning~\cite{agarwal_sensitivity_2021, setlur_is_2021}, or other approach-specific factors receive only limited attention.

The scope of the investigation has the strongest effect on how general the final results are. Although the majority of the papers focus on \textbf{multiple datasets and models}~\cite{bouthillier_unreproducible_2019, agarwal_sensitivity_2021, cioba_how_2022, ye_crossfit_2021, zhou_curse_2020, zhao_closer_2021, jundi_how_2022, schick_its_2021, liu_what_2022, zhao_calibrate_2021, zhang_active_2022, boquet_reproducibility_2019, zong2024foolvisionandlanguage, alzahrani2024benchmarks}, there are studies that consider only a \textbf{single dataset or a single model}~\cite{pham_problems_2021, setlur_is_2021, mccoy_berts_2020, bouthillier_accounting_2021, dodge_fine-tuning_2020, sellam_multiberts_2022}, limiting the generalisability of the results. \textbf{Only in specific cases is the investigation specifically designed to observe the effects of systematic choices}, such as the number of samples or model size, and \textbf{interactions between randomness factors}~\cite{reimers_reporting_2017, dehghani_benchmark_2021, zhong_are_2021, lu_fantastically_2022, efrat_lmentry_2022, weber-etal-2023-mind, chen-etal-2022-revisiting, reif-schwartz-2024-beyond, pecher2024sensitivity}. Finally, \textbf{only a minority of papers consider an out-of-distribution setting}~\cite{mccoy_berts_2020, zhou_curse_2020, cioba_how_2022, ye_crossfit_2021, zhang2023prompting, agrawal-etal-2023-context, zong2024foolvisionandlanguage} or \textbf{investigation across different languages}~\cite{zhang2023prompting, gan-mori-2023-sensitivity, zhang2024impact}.

\subsection{Investigation Methodology}

Overall, \textbf{the differences in how the investigation of the effects of different \textit{randomness factors} is done are small across different papers}, making the investigation quite consistent. In all papers, the investigation is performed by changing the \textit{randomness factor configuration} (such as changing the value of the random seed) and observing the changes in the behaviour of the model and its results. In the early papers, this was done by introducing the change randomly into the \textit{configuration}, without any control over the remaining randomness factors~\cite{reimers_reporting_2017, bouthillier_unreproducible_2019, cioba_how_2022, ye_crossfit_2021, mccoy_berts_2020, zhou_curse_2020, zhao_closer_2021, jundi_how_2022, zhong_are_2021, schick_its_2021, liu_what_2022, zhang_active_2022, bouthillier_accounting_2021, sellam_multiberts_2022, pezeshkpour2023large, reif-schwartz-2024-beyond}. However, the most common strategy is to exert additional control over the investigation by fixing the \textit{configuration} of the non-investigated randomness factors to a specific value while randomly varying the value for the single investigated factor~\cite{pham_problems_2021, lu_fantastically_2022, zhao_calibrate_2021, efrat_lmentry_2022, boquet_reproducibility_2019, summers_nondeterminism_2021, zhang2024impact, madaan2024quantifying, gundersen_reporting_2023, sclar_quantifying_2023, voronov2024mind}. In specific cases, the fixed investigation is repeated across different fixed values for the non-investigated factors and aggregated over them to better deal with interactions~\cite{webson-pavlick-2022-prompt, pecher2024sensitivity}. Finally, another investigation strategy changes the \textit{configuration} for multiple randomness factors at the same time, and their effects are disentangled at the end~\cite{dodge_fine-tuning_2020, agrawal-etal-2023-context, weber-etal-2023-mind, chen-etal-2022-revisiting, wei2024unveiling, zong2024foolvisionandlanguage}.

The choice of \textbf{how many \textit{randomness factors} to investigate at the same time also plays an important role} in the investigation methodology. Many papers consider only a single \textit{randomness factor} \cite{reimers_reporting_2017, bouthillier_unreproducible_2019, dehghani_benchmark_2021, schick_its_2021, liu_what_2022, pezeshkpour2023large, sclar_quantifying_2023}. However, such investigation can lead to biased results due to the interactions between different \textit{randomness factors}. Using multiple \textit{randomness factors} in a single investigation allows for a more detailed exploration of the behaviour of models in the presence of different sources of randomness.  Most approaches investigate multiple \textit{randomness factors}, although the number can be quite low (2-3)~\cite{cioba_how_2022, mccoy_berts_2020, zhou_curse_2020, zhao_closer_2021, jundi_how_2022, zhong_are_2021, efrat_lmentry_2022, zhang_active_2022, bouthillier_accounting_2021, boquet_reproducibility_2019, dodge_fine-tuning_2020, sellam_multiberts_2022, lu_fantastically_2022, zhao_calibrate_2021, zhang2024impact, chen-etal-2022-revisiting, reif-schwartz-2024-beyond, zong2024foolvisionandlanguage}. Comprehensive investigation of a large number of factors is considered only in specific cases~\cite{pham_problems_2021, summers_nondeterminism_2021, zhuang_randomness_2022, weber-etal-2023-mind, gundersen_reporting_2023, pecher2024sensitivity}.

Due to the potentially significant impact of the interactions between \textit{randomness factors} and systematic choices, it is important to define \textbf{how to consider these interactions and how to deal with them} to disentangle the effects of different \textit{randomness factors} and investigate \textit{how the systematic choices affect the findings}. Many papers do not use any specific way to distinguish between the effects of different \textit{randomness factors} \cite{reimers_reporting_2017, bouthillier_unreproducible_2019, dehghani_benchmark_2021, setlur_is_2021, cioba_how_2022, ye_crossfit_2021, mccoy_berts_2020, zhou_curse_2020, agarwal_sensitivity_2021, schick_its_2021, liu_what_2022}. As mentioned before, this may lead to the results being biased and can be traced as a direct cause of some of the open problems discussed in Section~\ref{sec:open-problems}. A simple way to consider the interactions is to use the fixed investigation~\cite{pham_problems_2021, zhao_closer_2021, lu_fantastically_2022, zhao_calibrate_2021, efrat_lmentry_2022, boquet_reproducibility_2019}. However, such an approach is limited in its investigatory strength, due to the randomness still being present in the choice of the fixed \textit{configuration}. A better way is to search through combinations of \textit{configurations} for all the \textit{factors} at the same time~\cite{jundi_how_2022, zhong_are_2021, zhang_active_2022, bouthillier_accounting_2021, dodge_fine-tuning_2020, sellam_multiberts_2022, webson-pavlick-2022-prompt, agrawal-etal-2023-context, voronov2024mind, pecher2024sensitivity}. However, \textbf{this significantly increases the computational cost of the whole investigation} \cite{bouthillier_accounting_2021, dehghani_benchmark_2021, zhuang_randomness_2022}. In case of systematic choices, the most straightforward approach is to repeat the investigation for different systematic choices, such as the number of samples or model size~\cite{zhang2024impact, zong2024foolvisionandlanguage, reif-schwartz-2024-beyond, wei2024unveiling, ma-etal-2023-large, chen-etal-2022-revisiting, pecher2024sensitivity}.

Another important aspect to consider is \textbf{how to obtain representative results from the investigation}. The representativeness of results mainly depends on how many \textit{randomness factor configurations} are explored, i.e., how many training and evaluation runs are used during the investigation. Many papers choose the number of different \textit{randomness factor configurations} only arbitrarily, often choosing lower number (3, 5, 40, 41, 86, 100, 1000) without any explicit explanation why the given number was chosen \cite{reimers_reporting_2017, bouthillier_unreproducible_2019, zhou_curse_2020, boquet_reproducibility_2019, ye_crossfit_2021, liu_what_2022, mccoy_berts_2020, zhao_closer_2021, schick_its_2021, pham_problems_2021, efrat_lmentry_2022, bouthillier_accounting_2021}. Similarly, when considering multiple \textit{randomness factors} and interactions between them, it is common to arbitrarily select some number (usually small) of \textit{configurations} for each factor and then determine the number of runs as all possible combination of these \textit{configurations} \cite{jundi_how_2022, zhong_are_2021, dodge_fine-tuning_2020, sellam_multiberts_2022, webson-pavlick-2022-prompt}. For example, when considering 3 different \textit{randomness factors}, with 5 different \textit{configurations} for each, the final number of runs is determined as 125 ($5^3$). In specific cases, all possible \textit{configurations} of a single \textit{factor} can be explored \cite{dehghani_benchmark_2021, agarwal_sensitivity_2021, lu_fantastically_2022, zhao_calibrate_2021, zhang_active_2022, voronov2024mind} or the number can be determined dynamically (e.g., until something happens or a threshold is overcome)~\cite{zhan2024unveiling, zong2024foolvisionandlanguage, pecher2024sensitivity}. However, this happens only when the number of possible \textit{configurations} is small or the added computation cost is not as significant.

Majority of the papers \textbf{evaluate the effects of randomness from multiple runs based on an aggregated value from these runs}, either as a single value of mean, standard deviation (with standard deviation not even being present in some papers), or number of failed runs \cite{bouthillier_unreproducible_2019, dehghani_benchmark_2021, pham_problems_2021, setlur_is_2021, cioba_how_2022, ye_crossfit_2021, mccoy_berts_2020, zhou_curse_2020, zhao_closer_2021, jundi_how_2022, zhong_are_2021, lu_fantastically_2022, schick_its_2021, zhao_calibrate_2021, efrat_lmentry_2022}, or as multiple aggregated values, such as average, worst and best performance, or the difference between them (e.g., difference between worst and best run, or difference between worst and average run) \cite{agarwal_sensitivity_2021, liu_what_2022, zhang_active_2022}. Some papers define new metrics, such as an average disagreement between models~\cite{summers_nondeterminism_2021, zhuang_randomness_2022}, a change in performance when models are ensembled~\cite{summers_nondeterminism_2021}, a relative gain~\cite{ajith2023instructeval, zhuang_randomness_2022}, or a factor importance~\cite{pecher2024sensitivity}. Only some papers evaluate the results of the investigation based on the distribution, mostly as a means for comparing different \textit{randomness factors}, models or the behaviour across datasets \cite{reimers_reporting_2017}. Some papers introduce special evaluation strategies based on statistical approaches that can better distinguish between effects of different \textit{factors} and deal with their interactions, and that can better estimate the results distribution from a small number of repeated runs to make the results more representative \cite{bouthillier_accounting_2021, boquet_reproducibility_2019, dodge_fine-tuning_2020, sellam_multiberts_2022}. Finally, only a minority of papers compare the learned representation instead of the predictions~\cite{summers_nondeterminism_2021, zhuang_randomness_2022, banerjee2024measuringmodelvariabilityusing}.

\subsection{Results from the Investigation}
\label{sec:investigation:results}

\textbf{Majority of the papers found that the effects of randomness have significant effects on the performance and stability across all approaches for dealing with limited data}, leading to differences in performance as high as 90\%~\cite{bouthillier_unreproducible_2019, dehghani_benchmark_2021, agarwal_sensitivity_2021, ye_crossfit_2021, mccoy_berts_2020, jundi_how_2022, zhong_are_2021, lu_fantastically_2022, schick_its_2021, zhao_calibrate_2021, efrat_lmentry_2022, zhang_active_2022, bouthillier_accounting_2021, boquet_reproducibility_2019, dodge_fine-tuning_2020, sellam_multiberts_2022}. Based on the randomness, the rankings in benchmarks and results of the comparison can change significantly~\cite{madaan2024quantifying, alzahrani2024benchmarks, zong2024foolvisionandlanguage}. Overall, the algorithmic factors were found to be more important than implementation, but only by a small margin~\cite{summers_nondeterminism_2021, zhuang_randomness_2022, gundersen_reporting_2023}. \textit{In-context} learning was found to be significantly sensitive to prompt format~\cite{weber-etal-2023-mind, voronov2024mind, sclar_quantifying_2023, zhan2024unveiling, zhang2023prompting}, choices of samples~\cite{zhong2023can, agrawal-etal-2023-context, zhang2023prompting} and order of choices in multi-choice QA~\cite{zong2024foolvisionandlanguage, wei2024unveiling}, leading to average performance difference of 30\% and up to 70\%. \textit{Fine-tuning} BERT multiple times and picking the best result, it is possible to outperform its more recent variants \cite{dodge_fine-tuning_2020}. The difference in performance between the worst performing and the best performing set of adaptation data in \textit{meta-learning} can be up to 90\% \cite{agarwal_sensitivity_2021}. On the other hand, the observed variance in other experiments was found to be lower, usually only a few percent difference (1\%-5\%), while still being enough to influence the results and comparisons between different models \cite{pham_problems_2021, zhao_closer_2021, liu_what_2022, reimers_reporting_2017}. However, \textbf{the effects of randomness are significantly higher in the out-of-distribution setting}~\cite{ye_crossfit_2021, mccoy_berts_2020, zhou_curse_2020, zhang2023prompting, agrawal-etal-2023-context, zong2024foolvisionandlanguage}.

The investigation results \textbf{indicate that the impact of interactions, systematic choices and experimental setup on the sensitivity to the effects of randomness and findings is significant in many cases}~\cite{gundersen_reporting_2023, zhang2024impact, madaan2024quantifying, zong2024foolvisionandlanguage, weber-etal-2023-mind, pecher2024sensitivity}. However, there is usually \textbf{low consensus} on how much the variance is reduced by different systematic choices. Some papers found that using more labelled samples or more training iterations reduces the stability ~\cite{agarwal_sensitivity_2021, jundi_how_2022, zhong2023can, agrawal-etal-2023-context, reif-schwartz-2024-beyond, chen-etal-2022-revisiting, pecher2024sensitivity}, although with diminishing returns \cite{zhao_closer_2021, ma-etal-2023-large}. On the other hand, other papers found that increasing the number of samples does not lead to significant improvement \cite{lu_fantastically_2022, zhang2023prompting, pezeshkpour2023large, reif-schwartz-2024-beyond, zong2024foolvisionandlanguage}, or even having negative effects on the stability \cite{zhao_calibrate_2021}. Similarly, some papers found that increasing the model size reduces the variance \cite{zhong_are_2021, zhang2024impact}, but others have found that it either has no effect or even worsens the situation \cite{efrat_lmentry_2022, chen-etal-2022-revisiting}. Optimising \textit{configuration} of specific randomness factors leads to lower sensitivity to other factors, such as using sophisticated sample selection reducing sensitivity to order~\cite{zhong2023can, agarwal_sensitivity_2021, zhang_active_2022}, although it varies across experimental setups~\cite{agarwal_sensitivity_2021, zhang2024impact} (e.g., using 16 random samples is equivalent to using 1 high-quality sample). Keeping the \textit{randomness factor configuration} fixed during the whole use of the model, such as using a single set of samples or defined order based on heuristics, can help reduce the variance \cite{setlur_is_2021}, but not in all the cases \cite{zhang_active_2022, boquet_reproducibility_2019}.  Finally, accounting for and investigating as many sources of randomness as possible helps with reducing variance at the cost of computational cost \cite{bouthillier_accounting_2021}.

The behaviour of \textit{randomness factors} is affected by the choice of models and datasets. \textbf{The inconsistency in results is present here as well.} Some papers indicate that there is different behaviour of the \textit{randomness factors} across different model architectures~\cite{reimers_reporting_2017, bouthillier_unreproducible_2019, dehghani_benchmark_2021, zhong_are_2021, lu_fantastically_2022, zhang2024impact, atreja2024prompt, madaan2024quantifying, wei2024unveiling} or datasets~\cite{zhou_curse_2020, ye_crossfit_2021, zhao_calibrate_2021, zhang2023prompting, zhang2024impact, atreja2024prompt, madaan2024quantifying}, with smaller datasets showing more variance \cite{dodge_fine-tuning_2020}. However, many of the results from investigation indicate that the behaviour of \textit{randomness factors} is consistent across all the model architectures, sizes and datasets \cite{agarwal_sensitivity_2021, mccoy_berts_2020, zhao_closer_2021, jundi_how_2022, schick_its_2021, liu_what_2022, efrat_lmentry_2022, zhang_active_2022, bouthillier_accounting_2021, boquet_reproducibility_2019, sellam_multiberts_2022, webson-pavlick-2022-prompt, sclar_quantifying_2023}.

The investigation of the effects of randomness often just determines that there is a sensitivity to a specific factor, but \textbf{without providing any additional analysis of how much it affects the models or what \textit{randomness factors} are most important}~\cite{setlur_is_2021, cioba_how_2022, ye_crossfit_2021, mccoy_berts_2020, zhou_curse_2020, schick_its_2021, liu_what_2022, zhang_active_2022, sellam_multiberts_2022, zhang2023prompting}. However, \textbf{the most interesting use for the results of the investigation is to provide more in-depth analysis}. For example, comparing between different randomness factors, datasets and models~\cite{pecher2024sensitivity}. Although both choice and order are significant contributors, some results indicate that choice is more important as with good samples the sensitivity to order is significantly lower~\cite{agrawal-etal-2023-context, ma-etal-2023-large, pecher2024sensitivity}. The prompt format can even affect the sensitivity to the chosen samples~\cite{zhang2024impact}. Other results indicate that random seeds, especially the ones used when \textit{fine-tuning}, are more significant than just the effects of implementation and hardware \cite{pham_problems_2021, zhong_are_2021, summers_nondeterminism_2021, zhuang_randomness_2022}, while others find the effects implementation and hardware to contribute more significantly \cite{boquet_reproducibility_2019}. The results from the investigation of almost all randomness factors at the same time indicate that the splitting of data has the most impact on the performance \cite{bouthillier_accounting_2021}. Finally, \textbf{analysing the results can better inform other choices later in the process}. For example, estimating the benefit of prompt formats based on closeness (similar to hyperparameter tuning) is not possible, as even the best and worst performing prompts can be close to each other~\cite{zhan2024unveiling}. In addition, the quality of the prompt is not dependent on how well the instructions are written, as even misleading or irrelevant instructions can lead to good results (e.g., better than the correct prompt)~\cite{webson-pavlick-2022-prompt}.

\subsection{Overview of Findings: Investigating Effects of Randomness}

The effects of randomness are \textbf{not investigated evenly across all groups of approaches for dealing with limited labelled data or modalities}. The investigation is more skewed towards randomness in language models, which goes hand in hand with the focus on natural language processing. However, the findings from \textit{meta-learning} and overall \textit{machine learning} approaches indicate that this mostly stems from the popularity of language model research and not from the impact the randomness has on the investigated approaches. The \textbf{effects of randomness are consistently significant across all approaches} for dealing with limited labelled data.

In addition, the \textbf{focus of the investigation is mostly on a few popular \textit{randomness factors}}, such as random seeds, model initialisation, sample choice and order, which may cause lower instability. Even though other \textit{randomness factors} receive minimal attention, it has no correlation with their importance. For example, \textbf{data split is considered the most important factor}, while being investigated only in a few papers. Another mostly ignored factors are the non-deterministic implementation and hardware, and the choice of which samples are labelled.

The \textbf{investigation strategy design is agnostic from data and the machine learning approach}, making its use consistent across all papers with only small modifications. However, these small modifications have a significant impact on the overall investigation, \textbf{often making it impossible to compare findings from different papers}. The number of training and evaluation runs are \textbf{chosen rather arbitrarily}, ranging from ones to thousands, \textbf{without any explanations behind the choice}. Almost no paper introduces heuristics for determining how many repeated runs are needed to estimate the underlying distribution of results. The only heuristic used is to explore all the possible \textit{randomness factor configurations} when their number is small, such as the permutations when the number of used samples is small ($< 10$).

The effects of randomness are usually \textbf{investigated only across a small number of \textit{randomness factors} at one time}. The number of considered \textit{factors} is usually 1-3, while other \textit{factors} may be mentioned, but not investigated. Only papers dealing with general machine learning cover a broader and more representative set of \textit{factors} at the same time. However, it is problematic to investigate multiple \textit{factors} at the same time due to the computational cost this introduces - the number of repeated runs raises exponentially with a number of \textit{randomness factors} \cite{bouthillier_unreproducible_2019, bouthillier_accounting_2021, dehghani_benchmark_2021}.

Even when evaluating multiple \textit{randomness factors} at the same time, \textbf{the interactions between them or the impact of systematic choices are not taken into consideration}. Many papers use the fixed \textit{randomness factor configuration} in the investigation, but it is still affected by the randomness. Only some papers use combinations of all \textit{randomness factor combinations}. However, this is always \textbf{done only for \textit{factors} that are considered in the paper} and not for all of them. In addition, no paper applies mitigation strategies to other \textit{randomness factors}.

\textbf{No analysis of the importance} of the \textit{randomness factors} is performed when investigating a single \textit{factor}. Similarly, when investigating multiple \textit{factors} at the same time, \textbf{almost no comparison is performed between these \textit{factors}}. This causes the investigation results to be limited in their extent as the problem is only recognised, but no further conclusions can be drawn from it. It is not possible to answer questions like: ``Are the effects of the \textit{factor} significant?'' or ``Which \textit{factor} is more important and should be focused on first?''.

The effects of randomness are evaluated in a multitude of ways. The most common approach is to use a simple aggregation over the runs (e.g., mean or standard deviation). However, as the experiments are performed on a limited number of training runs, the single aggregated value may not be representative of the underlying distribution. A \textbf{better approach for evaluating the effects is the use of statistical approaches} that allow to quantify the uncertainty still present in the models when only limited runs are used, or to determine the expected behaviour of the models in limit, while only having results from a significantly smaller number of runs. The evaluation is further complicated by the interactions between \textit{randomness factors} that can muddle the results and need to be taken into consideration using statistical approaches.

\textbf{Focus on the more specialised settings, such as out-of-distribution, multilingual, or using parameter-efficient fine-tuning methods, is severely limited}. However, few works that study these settings found the effects of randomness to be more significant in such settings, especially for out-of-distribution tasks or parameter-efficient methods.

A significant drawback regarding the results from the investigation is that \textbf{no obvious consensus exists regarding the behaviour of different \textit{randomness factors} across different models and datasets}. Many papers investigating the same setting come to contradictory findings about almost all parts of the investigation. Some papers observe a decrease in variance when the size of the model increases, while others found no obvious connection between the size of the model and the amount of observed variance (e.g., increasing size leads to an increase in variance). When it comes to datasets, some papers observed consistent results, with the same \textit{factor} being most significant across different datasets, while others observed different \textit{factors} being important on different datasets. This may be a result of disregarding interactions or using a sub-optimal experimental setup, as some papers observed both interactions and systematic choices have significant effects on the findings of investigation (e.g., the effect of one factor may be misattributed to another factor).

The most consistent finding from the investigation is that \textbf{increasing the number of samples reduces variance}. However, the decrease in stability has diminishing returns, where from a certain point, increasing available data has no more significant impact. This may also be due to the quality of samples used, as some papers found that 1 high-quality sample provides a reduction in sensitivity equivalent to 16 randomly selected samples.

\section{Determining the Origin of Randomness}
\label{sec:det_source_of_randomness}
In this task, the origin of randomness is determined, which represents a specific underlying problem that only manifests through the \textit{randomness factors}. The design of experiments is similar to the one when investigating the effects of randomness, but with a specific hypothesis that is formed and answered either empirically or using theoretical understanding of deep learning models (e.g., whether the label which is used as the last example in \textit{in-context learning} has an impact on the prediction, which would indicate recency bias). Out of 161 core papers covered by this survey, only a small fraction of 27 papers focus on this task, even though it is important for the mitigation of the sensitivity to randomness.

The design of experiments and further explanation of the behaviour is closely tied to the used \textit{machine learning approach} and the \textit{randomness factor} that was observed to cause the randomness.

\subsection{Origins of Randomness in Meta-Learning}

The focus on origins of randomness in \textit{meta-learning} is limited only to the adaptation data sampling \textit{randomness factor}~\cite{agarwal_sensitivity_2021}. The behaviour of different \textit{meta-learning} approaches is analysed using a combination of all the possible combinations of the adaptation samples from image datasets. The goal of the analysis is to determine the characteristics of the worst-case samples, i.e., the samples that cause the worst performance when used for adaptation. It is determined that these samples are not artefacts, as they are always correctly labelled, appear representative of the respective classes and are not visually different from other images. Based on this observation, it is speculated that the problem is due to the adversarial nature of the images. However, it is later disproved empirically as adversarial training does not reduce the effects of randomness, leaving the real reason why \textit{meta-learning} is sensitive to the choice of support samples open.

\subsection{Origins of Randomness in Language Model Fine-Tuning, Prompt-Based Learning and Parameter-Efficient Fine-Tuning}
As the main sensitivity and origins of randomness for \textit{language model fine-tuning}, general \textit{machine learning}, \textit{prompt-based learning} and \textit{parameter-efficient fine-tuning} are related to each other, we describe them together. The origin is analysed mainly using the random seed \textit{randomness factor} (overall across 14 papers), which affects the initialisation, order of samples and model randomness. Many different origins of randomness are identified. The experiments in this part are the most diverse (e.g., models, languages, factors) but often lead to similar origins of randomness.

Small availability of data that causes overfitting and catastrophic forgetting is identified as being one origin of randomness when fine-tuning language models \cite{zhao_closer_2021, hua_noise_2021}. This can cause the models to rely on shallow lexical features, even though the models were trained on abundant, feature-rich data beforehand \cite{zhao_closer_2021}. Another possible origin may be the bad fine-tuning start point in the transferred model, mainly in the higher layers (not only the newly initialised ones but also the pre-trained ones) that are sensitive to perturbations in the input.

However, catastrophic forgetting and small size of training data are disputed as causes by later papers \cite{mosbach_stability_2021, zhang_revisiting_2021, wu_zero-shot_2022}. Instead, it is only a symptom of another problem \cite{mosbach_stability_2021}. Looking at the failed fine-tuning runs (i.e., the runs that perform worse than random chance), it is identified that the language models suffer from vanishing gradients. As the cause of the vanishing gradients, a sub-optimal experimental setup is identified, such as using an optimiser without bias correction or a small number of training iterations~\cite{mosbach_stability_2021, zhang_revisiting_2021}. An additional cause of vanishing gradients is the sub-optimal initialisation point for the upper layers, which are specialised and cause significant variance when transferred \cite{zhang_revisiting_2021}.

Again, all these hypotheses are disputed \cite{wu_zero-shot_2022} and instead under-specification of the optimisation problem is proposed as the origin of the variance \cite{wu_zero-shot_2022, damour_underspecification_2020, mccoy_berts_2020}, especially in out-of-distribution datasets \cite{wu_zero-shot_2022, mccoy_berts_2020}. There exist multiple good solutions in the generalisation errors on the source dataset, but with wildly different generalisation errors on the target datasets, with the optimal solution lying in a non-flat region where even a small perturbation leads to large difference in error \cite{wu_zero-shot_2022, mccoy_berts_2020}. This makes the identically trained models encode different inductive biases \cite{damour_underspecification_2020} and makes the choice of minima rather arbitrary, being easily affected by small changes in initialisation and order of examples \cite{mccoy_berts_2020}. Under-specification was also identified as the main origin of randomness for prompt-based learning~\cite{koksal-etal-2023-meal} and parameter-efficient fine-tuning (mainly prompt-tuning)~\cite{chen-etal-2023-ptp}, which causes the loss surface to be remarkably steep and a small change in input causes a massive change in the loss space.

Furthermore, it is postulated that the large variance in performance is due to the strong inter-example correlations in datasets \cite{zhou_curse_2020}. When the prediction of a single sample changes, this correlation causes simultaneous change in a large proportion of the predictions in the dataset. This was observed when analysing the training trajectory in an experiment, where the model checkpoints at the same step of training from different initialisations had significant differences in performances. The individual predictions of each sample were highly unstable and changed constantly, which caused large fluctuations. However, no explanation for the unstable behaviour in individual samples is provided, besides being an inevitable consequence of current datasets \cite{zhou_curse_2020}.

Finally, it is claimed that randomness is the inevitable outcome of the stochastic training and a basic amount of instability will be always observed, equivalent to changing 1 bit in the weights~\cite{summers_nondeterminism_2021}.

\subsection{Origins of Randomness in Prompting and In-Context Learning}
The origins of randomness in \textit{in-context learning} are analysed across three \textit{randomness factors} (overall 13 papers): 1) order of data (both samples and answer choices in multi-choice question answering) \cite{zhao_calibrate_2021, lu_fantastically_2022}; 2) choice of samples \cite{liu_what_2022, zhang_active_2022}; and 3) prompt format.

The significant effects of different data orders, sample choices and prompt formats were found to be the consequence of highly unbalanced label distribution in the output \cite{zhao_calibrate_2021, lu_fantastically_2022, ma2023fairness, sclar_quantifying_2023, zhang2024batch}. The output distribution of the language models is influenced by the biases present in them \cite{zhao_calibrate_2021, fei-etal-2023-mitigating, reif-schwartz-2024-beyond, zhao-etal-2024-correcting-language, zong2024foolvisionandlanguage, pezeshkpour2023large, zheng_large_2023}. For the order of in-context examples, four biases were identified that cause these effects~\cite{zhao_calibrate_2021, fei-etal-2023-mitigating, reif-schwartz-2024-beyond, zhao-etal-2024-correcting-language}: 1) majority label bias, where the output is biased towards the more frequent answer in the prompt (e.g., class imbalance in prompt); 2) recency bias, where the model tends to output the labels that appear towards the end of the prompt; and 3) common token bias, where the model is biased to output more common tokens from the pre-training distribution; and 4) domain label bias~\cite{fei-etal-2023-mitigating} that is introduced by the domain texts. For order of answer choices, mainly two biases are identified~\cite{zong2024foolvisionandlanguage, pezeshkpour2023large, zheng_large_2023}: 1) positional bias, where the model prefers answers in specific positions; and 2) token bias, where model prefers specific answer symbol (such as Option A). The combination of these biases causes the model to have a highly unbalanced label distribution that needs to be calibrated to reduce the variance. In addition, the biases are enhanced by the choice of prompt format and in-context examples~\cite{reif-schwartz-2024-beyond, sclar_quantifying_2023, zhao-etal-2024-correcting-language}.

On the other hand, calibrating the distribution to reduce the variance and sensitivity to prompts was found to not be effective, as the bias still persists~\cite{lu_fantastically_2022, zong2024foolvisionandlanguage}. This leads to the hypothesis that the biases are not the sole cause of the highly unbalanced label distribution and the origin of randomness should be traced to something different, such as a combination of biases and relationships between the choices and distractor answers~\cite{zong2024foolvisionandlanguage}, or prior in-context examples~\cite{zhang2024batch}.

The random sampling of examples for the \textit{in-context learning} was determined to be the origin of randomness \cite{liu_what_2022, zhang_active_2022} that can also cause problems observed when using a different order of examples in prompts. Choosing the samples randomly, without taking similarity or other heuristics into consideration, may result in many low-quality demonstrations being selected that may contain spurious correlations and may not provide enough information \cite{zhang_active_2022, liu_what_2022}. When low-quality samples are used, different permutations and calibration of output distribution do not help reduce the variance, which disputes the claims in the previous works \cite{zhang_active_2022}. Choosing high-quality samples based on similarity or heuristics reduces the variance and allows different ordering and calibration to work \cite{zhang_active_2022, liu_what_2022}. Therefore, the real origin of randomness, which can also lead to unbalanced distribution and sensitivity to order, is the choice of low-quality, dissimilar samples \cite{zhang_active_2022, liu_what_2022}.

Finally, it was observed that sensitivity to prompt format is affected by how frequently the prompt appears in some variation in the data during training~\cite{gonen-etal-2023-demystifying}.

\subsection{Overview of Findings: Origins of Randomness}
The observed origins of randomness can be summarised as follows:
\begin{itemize}
    \item \textbf{Poor choice of samples} that are used for training, adaptation or in-context learning.
    \item \textbf{Overfitting} that causes \textbf{catastrophic forgetting}, and models to focus on shallow features.
    \item \textbf{Under-specification}, where multiple local minima with the same performance are present in training data, which are not consistent with testing data.
    \item \textbf{Highly unbalanced output label distribution} stemming from biases present in language models and incorrect ordering of samples.
    \item \textbf{Optimisation problems} caused by poor choices in the experimental setup, such as not using bias correction, training for a limited number of iterations or poor initialisation and re-use of some neural network layers.
    \item \textbf{Prompt format} mainly caused by biases, such as how often the specific prompt and words in it are encountered during training.
\end{itemize}

The individual \textbf{origins of randomness are closely tied to the \textit{machine learning approach} and the \textit{randomness factor}} investigated. This is mainly the consequence of the experiments that are designed to observe the behaviour of different approaches. Although many different origins are identified, all are tied to a small number of \textit{randomness factors}, specifically a choice of samples, an order of the samples in training, a prompt format and model initialisation.

The popularity of different \textit{machine learning approaches} and \textit{randomness factors} is clearly shown in this task for addressing the effects of randomness. The focus on determining the origins of randomness in \textit{meta-learning} is limited, while for \textit{prompting}, \textit{in-context learning} and their \textit{fine-tuning} it is most popular.

The \textbf{first effects of interactions between \textit{randomness factors} are observed}. In many cases, the different origins of randomness identified are later disputed and denoted as a simple consequence of other effects of randomness. For example, it is observed that the choice of samples in \textit{in-context learning} causes problems that are often incorrectly attributed to the ordering of these samples. In addition, different randomness factors were found to enhance the sensitivity, such as prompt format affecting the biases in models that are a source of the sensitivity to data order.

This further shows the \textbf{prevailing inconsistency of results} from different papers. The compounding effects between different \textit{randomness factors} and systematic choices cause significant problems when not properly addressed. Another aspect that plays a role is the simple evaluation, as the interactions and systematic choices are clearly shown when a more comprehensive evaluation is utilised.

\section{Mitigating the Randomness}
\label{sec:mitigation}

In the mitigation task, the effects of randomness are mitigated to reduce the variance as much as possible and to improve the overall performance using mitigation strategies. Mitigation represents the most popular task for addressing the effects of randomness (overall 112 papers). We divide these mitigation strategies into two separate groups: 1) \textbf{general} mitigation strategies; and 2) \textbf{problem-specific} mitigation strategies. 

The general mitigation strategies can be used to mitigate the effects of randomness stemming from any origin of randomness and \textit{randomness factor}. So far, the only way to deal with any origin of randomness is to use ensemble-like approaches (used in overall 13 papers), repeating the training and evaluation multiple times, each time with different \textit{randomness factor configuration}, and then aggregating the predictions (using majority voting or weighted average). The different versions of the ensemble strategy were shown to provide effective mitigation for sample choice~\cite{qin2023context}, model randomness and optimisation~\cite{fu2022worst, summers_nondeterminism_2021, wang-etal-2023-two, pecher2024fighting}, order of samples or answer choices~\cite{zong2024foolvisionandlanguage, pezeshkpour2023large}, the prompt format~\cite{allingham23simple, jiang2023calibrating, voronov2024mind} or data split~\cite{zheng_fewnlu_2022, mukherjee_clues_2021}. In some cases, it is the only available solution for mitigating the randomness, i.e., when the \textit{randomness factor} cannot be easily controlled, such as the effects of \textit{implementation of deep learning frameworks and hardware}. However, it introduces a significant increase to computation costs~\cite{bouthillier_accounting_2021}. Only some papers focus on reducing the cost of the ensemble strategy~\cite{pecher2024fighting, summers_nondeterminism_2021}.

On the other hand, the \textbf{problem-specific} mitigation strategies are designed to deal with a specific origin of randomness, making them dependent on the \textit{randomness factor} and the origins of randomness as analysed in Section~\ref{sec:det_source_of_randomness}, but also more popular (used in overall 99 papers). In addition, these approaches often provide benefits only for the specific setup (e.g., task type, dataset, or model).

\subsection{Poor Choice of Samples}

Addressing the sensitivity to the choice of samples is the most popular approach (overall 51 papers). Samples for different approaches are chosen using two approaches: 1) selecting representative samples for the whole process of training and evaluation; and 2) sampling data for the specific training and evaluation iteration. The first approach represents the real-world scenario where there is only a limited labelling budget available that needs to be effectively utilised by choosing samples with as much information as possible. For \textit{in-context learning} this approach is designed to select a single set of examples that perform well for all test samples, instead of selecting separately for each test sample. In the second approach, the samples are selected for a specific training iteration (for fine-tuning) or for a specific test example (for \textit{in-context learning}).

\subsubsection{Selecting Representative Samples}

When selecting a small number of representative samples ($< 100$), strategies that work with unlabelled data are often used. This includes the use of a simple clustering~\cite{chang_training_2021, yu-etal-2023-cold, cho2023prompt, koksal-etal-2023-meal, wu2023scattershot}. To choose \textit{K} samples, the available data are clustered based on their similarity~\cite{chang_training_2021, cho2023prompt}, diversity~\cite{koksal-etal-2023-meal, wu2023scattershot} or the pseudo-labels generated by a large language model~\cite{yu-etal-2023-cold} into \textit{K} clusters. Afterwards, samples are selected from each cluster either as the closest to the centre~\cite{chang_training_2021, cho2023prompt}, or based on weighting using uncertainty~\cite{yu-etal-2023-cold} or KL divergence~\cite{koksal-etal-2023-meal}.

Another possibility is to use active learning strategies to select the samples based on uncertainty, diversity or entropy~\cite{xu2024misconfidence, margatina-etal-2023-active}, or their combination with other properties such as complexity or quality~\cite{liu_what_2023, mavromatis2023examples}. Which combination of properties to consider can also be determined dynamically based on the approach used~\cite{pecher2024automatic}. The submodular functions are often used to optimise multiple properties at the same time~\cite{ji-etal-2024-submodular-based, qian2024subsastrengthenincontextlearning}. Another possibility to select the samples is to use reinforcement learning, similarly to active learning \cite{zhang_active_2022}.

A popular strategy is to select samples based on their quality. To determine the quality of samples, one solution is to train a linear regression model that determines the gain in performance when the sample with given characteristics is included in the process~\cite{jundi_how_2022, chang-jia-2023-data, nguyen2023context, rubin-etal-2022-learning}. Another possibility is to prompt a large language model to rate each sample based on how it contributes~\cite{liu-wang-2023-towards, li2023large}.

Another set of strategies selects samples in multiple steps. In each step, a separate sample property is optimised, such as first choosing the most diverse or informative samples and then refining them using their similarity to test sample or their quality~\cite{su_selective_2022, li-qiu-2023-finding, yang-etal-2023-representative, ye2023compositional}. To achieve this, determinantal point processes are often used~\cite{yang-etal-2023-representative, ye2023compositional}, a large language model is prompted to provide quality/informativeness scores for the samples~\cite{li-qiu-2023-finding}, or a graph-based approach is used~\cite{su_selective_2022}. 

Finally, specific mitigation strategies focus on selecting the samples, their order and the format of the prompt at the same time, using neural bandits~\cite{wu2024prompt}.

\subsubsection{Sampling Data for Training Iteration}

When selecting in-context examples for a specific training or evaluation iteration, the most common approach is to use the similarity to test sample~\cite{agrawal-etal-2023-context, chen-etal-2023-stabilized, an-etal-2023-skill, liu_what_2022}. The similarity selection is often combined with other properties, such as text characteristics~\cite{wu-etal-2023-self, zhang2023prompting}, entropy~\cite{peng2024revisiting, adiga2024designing} or margin~\cite{zhu-etal-2024-towards-robust-context}. Besides using different properties, a better representation can also be used, such as using large language models to transform each sample into how well it represents a certain skill~\cite{an-etal-2023-skill}.

Another possibility is to use adversarial training on the worst-case adaptation samples (found using greedy search) in \textit{meta-learning} \cite{agarwal_sensitivity_2021}. However, it provides no significant improvement in regards to stability for specific \textit{meta-learning} approaches \cite{agarwal_sensitivity_2021}.

\subsection{Overfitting and Catastrophic Forgetting}

How the overfitting and catastrophic forgetting is mitigated is closely tied to the approach for which it is designed. Therefore, we provide further categorisation based on these approaches.

\subsubsection{Overfitting in Meta-Learning}

One solution to reduce the overfitting in meta-learning approaches is the augmentation of the support and query data, and the task construction \cite{osahor_ortho-shot_2022}. Applying such augmentation increases the diversity of the samples and tasks, which improves the overall generalisation and reduces the sensitivity to the randomness~\cite{osahor_ortho-shot_2022}.

Another solution is the use of variance reduction algorithms in the optimisation-based meta-learning \cite{wang_variance-reduced_2021, yang_efficient_2022}. One possibility is to focus on first-order meta-learning algorithms and introduce the variance reduction only to them \cite{wang_variance-reduced_2021}. The variance reduction term is introduced to the gradient estimator in the task adaptation stage, motivated by the recursive momentum technique in \cite{cutkosky2019storm}. This term is initialised by averaging the gradients from randomly sampled tasks using the initial parameters and then is further updated using a weighted sum of mini-batch stochastic gradients across all sampled tasks \cite{wang_variance-reduced_2021}. As this foregoes the benefits of the bi-level meta-learning process, other papers propose to modify the variance reduction term (STORM \cite{cutkosky2019storm,levy2021storm+}) for use on bi-level meta-learning optimisation, in combination with a large number of training steps to reduce the amount of storage required \cite{yang_efficient_2022}. To achieve this, the variance-reduced gradient estimate is computed using stochastic gradients from two successive iterates while evaluating the gradient using the previous two iterates on the current batch of samples \cite{yang_efficient_2022}.

\subsubsection{Overfitting in Language Model Fine-Tuning}
The augmentation technique, in combination with regularisation, is also proposed when fine-tuning language models \cite{meng_tuning_2022}. The pre-trained language model is used as a generator for new data samples used for augmentation. To guarantee that samples with discriminative labels are generated, a meta-weighted maximum likelihood objective can be used for the tuning of the generator. As the resulting labels still contain some level of noise, a noise-robust procedure for training the final model is proposed as regularisation using label smoothing and temporal ensembling \cite{meng_tuning_2022}.

Another solution is to look at the disconnection of the problem statement between the pre-training and fine-tuning stages. When the objective or even the input and output distributions are different, the fine-tuning process can often be brittle, especially when only a small amount of data is available \cite{phang_sentence_2019}. A second pre-training step on data-rich supervised tasks can be used to better prepare the model for the fine-tuning stage, mitigating the brittleness and tendency to overfit on data \cite{phang_sentence_2019}.

The last set of approaches utilises regularisation to prevent overfitting. Injecting a Gaussian-like noise vector into the input, either before or during training, can be used as a stability regularisation, similar to augmentation \cite{hua_noise_2021, wu-etal-2022-noisytune}. Another regularisation technique is to perform fine-tuning updates only on a sub-network instead of the whole network \cite{zhang2018mixup, xu_raise_2021, zhang_fine-tuning_2022}. Instead of random selection, CHILD-TUNING, introduced in \cite{xu_raise_2021}, uses Fisher Information to determine the importance of parameters based on all samples before training. Afterwards, an unchanged sub-network with the most important parameters is selected for the update. The previous two approaches can be improved using Dynamic Parameter Selection, introduced in \cite{zhang_fine-tuning_2022}, which adaptively selects a promising sub-network to perform the fine-tuning steps on. The sub-network is chosen dynamically, i.e., different sub-network is potentially fine-tuned at each training step, based on the importance of the parameters. It greatly reduced the heavy computation costs introduced by CHILD-TUNING \cite{xu_raise_2021} that separate the process of fine-tuning and the decision of which sub-network to optimise. In addition, it can better capture the task-relevant parameters across sub-networks of different sizes and bring stable improvement in various parts of the model. However, when faced with limited resources, the previous approaches cannot select optimal sub-network~\cite{somayajula-etal-2024-generalizable}. As such, the previous approach is modified to represent the weights as a weighted mixup of pre-trained weights and the optimised weights, which is guided and optimised using attention.

\subsection{Under-Specification}

To deal with under-specification, models with stronger inductive biases that can help distinguish between many local minima should be proposed \cite{mccoy_berts_2020, zhou_curse_2020}, or the training sets should be better designed \cite{mccoy_berts_2020, zhou_curse_2020}. The better sets should represent the phenomena appearing in texts to reduce the probability of local minima not transferring to testing data \cite{mccoy_berts_2020}. In addition, the tasks used for large-scale pre-training should adhere to structure and level of compositionality (i.e., solving a combination of multiple tasks leading to the solution for a different, more complex task), and more diverse datasets in terms of syntax and lexicon should be constructed \cite{zhou_curse_2020}.

Another solution for dealing with under-specification is to utilise an ensemble of models \cite{hidey_reducing_2022, khurana_how_2021}. One possibility to construct the ensemble is to combine predictions from multiple trained models using their uniform average \cite{hidey_reducing_2022}. Another possibility is to use stochastic weight averaging (SWA) as regularisation for the ensembles \cite{khurana_how_2021}. Instead of training the model multiple times, SWA creates the ensemble by averaging over different snapshots of model weights. Such an ensemble explores other solutions around the local minima, while also reducing computation cost as the ensembling at weight level needs only a single training run \cite{khurana_how_2021}. Although the SWA method consistently achieves better stability of performance than standard fine-tuning when used over different values of random seeds, the increase in stability is only slight \cite{khurana_how_2021}. In addition to ensembling, distillation and co-distillation can also be used \cite{hidey_reducing_2022}. 

To deal with the under-specification in prompt-based learning and prompt-tuning, the best approach is to train over different prompts at the same time~\cite{koksal-etal-2023-meal}, creating a kind of ensemble. Another well-performing approach is to train using noise regularisation, adding noise to the input text and embeddings~\cite{chen-etal-2023-ptp}.

\subsection{Highly Unbalanced Label Distribution}

To deal with highly unbalanced label distribution in the \textit{in-context learning} it is proposed to use calibration. One possibility is to use contextual calibration, where the output distribution is first estimated using neutral prompts \cite{zhao_calibrate_2021, fei-etal-2023-mitigating, zhou_batch_2023, liu-wang-2023-towards}. The neutral prompts contain only in-context samples with the testing sample having its text replaced with words without semantic meaning, such as ``N/A''. Using multiple such prompts, the distribution between labels can be estimated and then corrected using affine transformation \cite{zhao_calibrate_2021}. For example, if the model outputs positive labels in more than 50\% of cases (e.g., in 65\%) in binary classification, the testing example is considered to be positive only when the confidence is higher than this threshold (e.g., more than 65\%). Besides words without semantic meaning, other studies observed that sampling from the text corpus~\cite{fei-etal-2023-mitigating} or the specific batch~\cite{zhou_batch_2023} leads to better estimation of the distribution. 

Instead of using neutral prompts, it is possible to parameterise the distribution using Bayes rule \cite{min_noisy_2022}. The language model is not presented with \textit{K} concatenated samples in a single prompt but instead using one sample at a time. Obtaining output probabilities and multiplying them together can be used to estimate the distribution, while also reducing the dependency on the ordering of samples \cite{min_noisy_2022}. Compared to direct distribution estimation, this provides larger improvements in terms of stability when there are fewer training samples, the data is imbalanced, the number of classes is large, or a generalisation to unseen labels is required \cite{min_noisy_2022}.

However, other studies found that previous calibration strategies do not lead to significant mitigation of the biases~\cite{reif-schwartz-2024-beyond}. Instead, they propose to do a leave-one-out calibration using the in-context examples, where the probability is estimated by leaving out the specific in-context examples~\cite{reif-schwartz-2024-beyond}. Adding noise to the sample parameters was found to be another effective calibration technique~\cite{zhao2024noisyicl}.

The calibration approaches are used to deal with the sensitivity to the order of choices in multi-choice question answering as well~\cite{zheng_large_2023, wei2024unveiling, zong2024foolvisionandlanguage}. However, calibration was again found to be sub-optimal, leading only to negligible improvements~\cite{zong2024foolvisionandlanguage}. The best-performing strategy was ensemble, where the evaluation is run multiple times with different orderings, and then using majority voting~\cite{pezeshkpour2023large, zong2024foolvisionandlanguage}, or fine-tuning the model further using different choice orders~\cite{zong2024foolvisionandlanguage}.

Apart from calibration, the highly unbalanced distribution is addressed by finding the best ordering of examples \cite{lu_fantastically_2022, kumar_reordering_2021, xu2024context, liu2024let}. One possible approach to automatically generate good prompt orderings is to use heuristics and entropy \cite{lu_fantastically_2022} or KL divergence~\cite{xu2024context}. Another possibility is to search over possible permutations of training samples to find the best performing one using genetic algorithm \cite{kumar_reordering_2021}. Instead of greedy-search, curriculum learning can be used as well, to order the in-context examples from easiest to hardest~\cite{liu2024let}.

Finally, a simple solution is to just use longer context and more samples, as it leads to lower sensitivity to order~\cite{bertsch2024context}.

\subsection{Optimisation Problems}
The mitigation strategies for dealing with optimisation problems in \textit{language model fine-tuning} are simply specific fixes for the poor choices that introduce the effects of randomness. One of the fixes is to increase the number of iterations considerably and introduce a bias correction and small learning rate to avoid vanishing gradients \cite{mosbach_stability_2021, zhang_revisiting_2021}. 

Another solution is to modify the way the additional layers are initialised \cite{dauphin_metainit_2019}. Instead of initialising the layers randomly, which can lead to bad initialisation, a meta-learning approach can be used to find an initialisation that is amenable to learning using gradient descent \cite{dauphin_metainit_2019}. In addition, badly performing initialisation can be detected as early as the second training iteration thanks to the high correlation of performance at the start and the end of the fine-tuning process \cite{dodge_fine-tuning_2020}. Starting many runs with different initialisations and stopping the ones with bad performance in the early stages of fine-tuning, it is possible to search through the whole space of random initialisation to find the best performing ones without any significant increase in computation cost \cite{dodge_fine-tuning_2020}. 

Yet another solution is to also randomly re-initialise the top layers of the language models, in addition to the classification layer \cite{zhang_revisiting_2021}. The number of re-initialised layers depends on the problem statement and the model, with more complex models and datasets benefiting from a larger number of re-initialised layers, but helps only up to a certain point \cite{zhang_revisiting_2021}.

\subsection{Prompt Format}

Another popular focus is on mitigating the sensitivity to prompt formats (overall 32 papers). One strategy for dealing with the prompt format sensitivity is automatically selecting good performing prompts~\cite{shum-etal-2023-automatic, gonen-etal-2023-demystifying, zhang-etal-2023-auto, sorensen-etal-2022-information}. Starting from a seed prompt format, new prompts are automatically generated as a paraphrase using large language models or using back-translation. The generated prompts are ranked based on how well they perform on development set~\cite{shum-etal-2023-automatic}, training a separate ranking model~\cite{zhang-etal-2023-auto}, or using heuristics such as perplexity~\cite{gonen-etal-2023-demystifying} or mutual information~\cite{sorensen-etal-2022-information}.

Besides generating new prompts, another possibility is to optimise the seed prompt directly. To achieve this, the influential words in the prompt are identified and then iteratively replaced using unmasking~\cite{zhan2024unveiling}, starting from the most influential ones. Instead of replacing the words one at a time, it is possible to use a large language model to rate the prompt on a specific mini-batch and suggest changes to it to achieve lower sensitivity~\cite{pryzant-etal-2023-automatic}. 

When optimising the prompt, a popular strategy is to add soft prompt embeddings that are then further optimised~\cite{sun_evaluating_2023}. However, using soft prompts was observed to be significantly sensitive to their initialisation. To deal with this sensitivity, a meta-learning approach is utilised to find a good initialisation~\cite{qin-etal-2023-learning, hou-etal-2022-metaprompting, pan-etal-2023-self}. Besides meta-learning, the initial prompt can be improved by training it on a few source tasks and then fine-tuning it on the target task~\cite{vu-etal-2022-spot}, or optimising its representation using a separate neural network~\cite{razdaibiedina-etal-2023-residual, liu2024stablept}.

A popular approach is to use multiple prompt formats at the same time. Similarly to the previous approaches, the seed prompt is used to generate new ones, e.g., by paraphrasing the hard ones. These new prompts are then used to fine-tune the models on all of them at the same time~\cite{chen-etal-2023-stabilized, sanh2022multitask, li2023mixpro, zhou-etal-2022-prompt}. Another possibility is to use these prompts in an ensemble, where each prompt is used to obtain a prediction and the final evaluation is done as a majority voting or (weighted) average of the partial predictions~\cite{allingham23simple, jiang2023calibrating, voronov2024mind}.

\subsection{Overview of Findings: Mitigating Origins of Randomness}

Mitigation strategies already receive the majority of the focus from the tasks for addressing the effects of randomness (112 papers). Overall, the main focus of mitigation is dedicated to \textit{prompting} approaches (including \textit{in-context learning} and \textit{prompt-based learning}) (83 papers) and their sensitivity to prompt format, choice of samples and the order of samples and answer choices. 

The proposed \textbf{mitigation strategies are closely tied to the origin of the randomness}. As they are explicitly designed to deal with the specific origin of randomness and the \textit{randomness factor} it was observed around, the proposed strategies are unusable for other \textit{randomness factors} and origins of randomness. The \textbf{best-performing mitigation strategies are ensembling and further model fine-tuning}, shown to provide benefit for many different randomness factors. \textbf{Some \textit{randomness factors} can be mitigated only using the ensemble strategy}, such as the non-deterministic \textit{implementation of deep learning frameworks and hardware}. As such, it is an ideal mitigation strategy, when the introduced computation costs are not a problem. Specific mitigation strategies should be used when only a specific factor needs to be mitigated while minimising the cost of the mitigation.

Even though the mitigation strategies address the specific origin of randomness, \textbf{many papers do not provide any kind of extensive analysis of where the randomness originates from}. In some cases, the works reference other papers that found this origin, but \textbf{in many cases no explanation is provided as to why the mitigation strategy targets the specific origin of randomness}.

\textbf{Each mitigation strategy incurs additional computation cost}. The most significant increase is introduced by the ensemble mitigation strategy, as it requires a large number of model training and evaluation runs (the mitigation effectiveness depends on the number of models in the ensemble). The expected computation cost is several times higher than in mitigation strategies that require only a single training run. So far, no general mitigation strategy exists that could provide mitigation without high computational costs. In the case of problem-specific mitigation strategies, the computation cost depends on multiple aspects, such as how often it is applied (once before the training run or at each training step), how expensive the approach used (genetic algorithm or simple clustering), or how many repeated runs are needed (e.g., when ensembling is used).

\textbf{Rffectiveness of the mitigation strategies is evaluated in a simple fashion} by comparing the performance metric of the model after applying the mitigation strategy. The comparison is usually done on a single aggregated value, often with a low number of repeated runs (sometimes even one), while foregoing the reporting of standard deviation or confidence intervals. In addition, the \textbf{interactions between \textit{factors} and systematic choices are often completely disregarded}.

Such a simple evaluation that disregards the interactions and systematic choices may be the \textbf{main reason for inconsistency in results}. Similar to the \textit{investigate} (Section~\ref{sec:investigation}) and \textit{determine} (Section~\ref{sec:det_source_of_randomness}) tasks, multiple papers report contradictory findings. The mitigation strategy that works in one paper provides no benefits in another. 

Finally, \textbf{only a single mitigation strategy is often applied at one time in most papers}. Only a few papers use multiple strategies and compare their benefits.

\section{Comparing and Benchmarking: The Impact of Randomness in Low Data Regimes}
\label{sec:benchmark-comparison}

In this task, the effects of randomness are taken into consideration when comparing and benchmarking different approaches for learning with limited labelled data. As the effects of randomness can have a significant effect when comparing, or when designing benchmarks, modifications to the process of benchmarking and comparing are required. This is recognised by the researchers, as the number of papers focusing on these tasks is growing (overall 33 out of 161 papers). Even though the benchmarks and comparisons between approaches are closely related, the proposed modifications that can deal with the effects of randomness look at the problem from different perspectives. 

When simply \textbf{comparing} the approaches, the modifications are introduced at the end of training. The design of the modified comparisons is independent of the training process and, in theory, can be applied to any dataset, model, or even benchmarks. The independence from the training process makes the modifications easier to design,but often requires specific assumptions, such as performing multiple runs. The modifications are usually designed to be simple, often taking the form of best practices for the comparisons (e.g., use statistical significance tests on the distribution of results). An important aspect of the comparisons is what metrics are reported.

On the other hand, the \textbf{benchmarks} are specifically designed to take the randomness into consideration from the start of training. The whole training process is modified based on desiderata that address the randomness, making benchmarks closely tied to the training process. Even though it makes the benchmarks harder to design, they often provide better explanations and comparative power as no assumptions and constraints are needed.

\subsection{Comparing Approaches in Low Data Regimes}

Comparisons between different approaches in low data regimes are significantly affected by the effects of randomness~\cite{pecher2024fine}. Changing the \textit{randomness factor configuration}, the ranking of models can vary significantly (as shown in Section~\ref{sec:investigation:results})~\cite{dehghani_benchmark_2021, dodge_fine-tuning_2020, reimers_reporting_2017, mizrahi2023state, alzahrani2024benchmarks, gupta2024changinganswerorderdecrease}. In addition, when using only a small number of runs, there is a high possibility of cherry-picking results that confirm our hypothesis~\cite{gundersen_reporting_2023}.

To deal with the effects of randomness, many papers propose to compare on multiple evaluation runs to achieve unbiased comparisons~\cite{schick_its_2021, reimers_reporting_2017, bouthillier_unreproducible_2019, dehghani_benchmark_2021, sellam_multiberts_2022, bouthillier_accounting_2021, dror_deep_2019}. Especially for prompt formats, where it was observed that one format is not ideal for different models~\cite{stefanik-etal-2023-resources, sclar_quantifying_2023, voronov2024mind}, it is important to compare models across multiple runs~\cite{stefanik-etal-2023-resources, sclar_quantifying_2023, voronov2024mind, mosbach-etal-2023-shot, mizrahi2023state, alzahrani2024benchmarks, gupta2024changinganswerorderdecrease}, keeping as much of the setup (e.g., size of models), using either the distribution of results or the best performance.

Comparing based on score distribution across multiple runs allows us to perform comparisons using statistical significance testing, which is a simple but powerful modification for the comparisons~\cite{reimers_reporting_2017, bouthillier_unreproducible_2019, dehghani_benchmark_2021, sellam_multiberts_2022, bouthillier_accounting_2021, dror_deep_2019, madaan2024quantifying}. Drawing conclusions from such comparisons reduces the bias and the risk of rejecting promising approaches or falsely accepting weaker ones \cite{reimers_reporting_2017, gundersen_reporting_2023, schick_its_2021}.

However, comparing between score distributions may not be as straightforward in all cases \cite{bouthillier_accounting_2021, sellam_multiberts_2022, dror_deep_2019}. To provide truly unbiased comparisons, as many sources of randomness (i.e., \textit{randomness factors}) as possible should be randomised \cite{reimers_reporting_2017, bouthillier_accounting_2021}. This not only leads to a better evaluation of the variance associated with each source of randomness but the error of the expected performance is also reduced. However, such comparisons start to become problematic when the evaluated models are not under our control (i.e., behind API) and the only controllable variance is the data we use. In addition, accounting for all sources of randomness is often not feasible due to the computational cost of running training multiple times \cite{bouthillier_accounting_2021}. To reduce the cost, a statistical bootstrap approach can be used to estimate the behaviour pm different settings from lower number of runs~\cite{sellam_multiberts_2022, polo2024efficient}.

Besides simple metrics, multiple papers propose to report additional ones that can allow for more unbiased comparisons using statistical tests. Such metrics include standard deviation, maximum, minimum, mean and confidence intervals \cite{schick_its_2021, bragg_flex_2021}, the score distributions and individual episode results from multiple runs, data splits, prompt formats or even across different datasets \cite{dehghani_benchmark_2021, dror_deep_2019, zheng_fewnlu_2022, mukherjee_clues_2021, bragg_flex_2021, sclar_quantifying_2023, mizrahi2023state}. Last, but not least, reporting how the ranking of the model changes when introduces a small change was found to be important for large language models\cite{alzahrani2024benchmarks, mizrahi2023state}.

\subsection{Benchmarks Taking Randomness Into Consideration}

Multiple benchmarks designed for evaluating \textit{few-shot learning} are starting to take the problem of randomness and its effects on the stability into consideration \cite{bragg_flex_2021, mukherjee_clues_2021, setlur_two_2021, ye_crossfit_2021, zheng_fewnlu_2022, efrat_lmentry_2022}. However, the extent to which the randomness is addressed is different across the benchmarks.

In the most simple case, only some of the properties related to stability and/or randomness are included in the design of the benchmark \cite{bragg_flex_2021, mukherjee_clues_2021, setlur_two_2021, ye_crossfit_2021}. This includes focus on variable number of shots and classes in \textit{few-shot learning} to better simulate real-world scenarios and to test robustness of models to these systematic changes \cite{bragg_flex_2021} or comparing the performance from multiple runs with the human performance \cite{mukherjee_clues_2021} to allow for paired statistical tests \cite{bragg_flex_2021}. Another modification is providing means to evaluate both in-distribution, as well as out-of-distribution \cite{setlur_two_2021}, or looking at the behaviour of models in cross-task settings \cite{ye_crossfit_2021}.

Later benchmarks also explicitly encourage the evaluation on multiple training runs \cite{mukherjee_clues_2021, ye_crossfit_2021, zheng_fewnlu_2022}. Such benchmarks mostly provide multiple splits of data for training and evaluation \cite{mukherjee_clues_2021, ye_crossfit_2021, zheng_fewnlu_2022}. To explore the behaviour in cross-task settings, the splitting is also done on the level of tasks, with some tasks being used only for testing purposes \cite{ye_crossfit_2021}.

Finally, some benchmarks are specifically designed to determine the sensitivity of the approaches to the effects of small, random changes~\cite{efrat_lmentry_2022, xu-etal-2023-multiinstruct, santurkar_whose_2023, ajith2023instructeval, liang2023holistic, zhu2023promptbench}. One benchmark focuses on simpler tasks that can be quickly and automatically evaluated, and are trivial for humans (i.e., tasks that elementary school students should perform perfectly at)~\cite{efrat_lmentry_2022}. For such tasks, the sensitivity to effects of randomness can be more easily and directly estimated and a sensitivity score can be estimated that is used to explicitly penalise models for their brittleness~\cite{efrat_lmentry_2022}. For more complex tasks, the evaluation is performed across different small perturbations (mainly to prompt format) and a specific sensitivity metric is designed, such as the spread of performance over these changes or the mean relative gain of the model~\cite{xu-etal-2023-multiinstruct, santurkar_whose_2023, ajith2023instructeval, liang2023holistic, zhu2023promptbench}.

\subsection{Overview of Findings: Comparisons and Benchmarks}
\label{sec:benchmark-findings}

The comparisons are specifically designed to deal with the significant effects of randomness when dealing with limited labels. \textbf{Changing the randomness factor configuration, such as using different prompt formats, leads to significant changes in rankings.} Even when comparing across multiple training runs, it was observed that \textbf{using a single aggregated value leads to biased comparisons}. The most common modification is to compare between result distributions using statistical tests instead. Although this leads to more unbiased comparisons, there are still problems present. In the case of approaches that use prompts, the best solution is to optimise the prompt format for each model (as the specific format is ideal only for specific models), reporting either the spread across different formats or the best performance from the different runs.

Such comparisons are sensitive to the number of training runs, as the \textbf{distribution of results can be objectively determined only with a larger number of runs}. However, running multiple training runs was found to be infeasible in many papers. To deal with this problem, approaches that approximate the distribution without the need for many training runs were proposed recently, such as statistical bootstrapping.

The modified comparisons are utilised only for comparing between different approaches across datasets and tasks. However, it is \textbf{important to also compare between effects of \textit{randomness factors}}, while taking the interactions into consideration. Even though not specifically designed for it, the modifications should be applicable to this problem as well.

When it comes to benchmarks, the randomness is only starting to be taken into consideration, leading to very simple ways to address it. \textbf{Most benchmarks simply evaluate across multiple splits, training runs or prompt formats, while keeping the number of runs quite small}, or simply explore some properties related to stability. In addition, even though multiple training runs are performed, many benchmarks compare the models only on a single aggregated value. \textbf{Only few benchmarks explicitly measures the sensitivity of models to effects of randomness} and use it to penalise the final performance metric.

\section{Challenges, Open Problems and Next Steps}
\label{sec:open-problems}

Based on the findings from the comprehensive literature overview presented in the previous chapters, we identify seven challenges and open problems for addressing the effects of randomness on the stability of learning with limited labelled data.

\subsection{Focus on Small Fragment of Randomness}

The scope of investigating the effects of randomness, determining its origin and the mitigation of these effects is limited in its extent. The problem can be observed on multiple levels:
\begin{itemize}
    \item \textit{Approaches and modality} - the focus is mostly on the approaches utilising language models (\textit{fine-tuning} and \textit{in-context learning}), while focus on other machine learning approaches, such as \textit{meta-learning} or even using \textit{parameter-efficient fine-tuning} methods, is severely limited in its extent. The majority of the focus is on natural language processing, as it is the modality used in language models, limiting the analysis of behaviour on other modalities. Therefore the effects of randomness are addressed only on a specific sub-part of learning with limited labelled data, even though the randomness also has significant effects on other machine learning approaches (investigated in this survey).
    \item \textit{Randomness factors} - many \textit{factors} receive limited attention, such as the effects from splitting data, choosing what data is labelled, model randomness, or the impact of non-deterministic implementation and hardware. Instead, the focus is on \textit{factors} in later stages of the training process, such as data order, prompt format or choice of data. This limits the analysis of randomness to a few popular \textit{randomness factors}. The under-researched \textit{factors} may still be important, but there is no consensus due to the limited focus. Some papers claim the effects of these \textit{factors} to be insignificant \cite{bouthillier_unreproducible_2019}, while others consider them to be the most significant contributors of variance \cite{bouthillier_accounting_2021, zheng_fewnlu_2022, mukherjee_clues_2021}.
    \item \textit{Number of randomness factors} - the experiments are designed to jointly analyse only a small number of \textit{factors} (1-3). Such experiments compound the limited focus on some \textit{factors} and may leave the results of the investigation vulnerable to unaddressed effects of other sources of randomness that may undesirably skew the results.
    \item \textit{Settings and tasks} - the experiments mainly consider in-distribution and monolingual settings for classification. Only a few papers deal with more specialised settings, such as out-of-distribution, multilingual, use of parameter-efficient fine-tuning methods, or more specialised tasks (summarisation, multi-choice question answering).
\end{itemize}

We argue that the scope should be widened across all tasks for addressing the effects of randomness, to better explore the behaviour of different \textit{machine learning approaches} across different modalities and \textit{randomness factors}. A simple extension would be to perform multiple experiments, each for different \textit{randomness factor} across all the different approaches the factor is relevant for. A more in-depth analysis would be to analyse the effects of randomness across all \textit{factors} in a single experiment. However, increasing the number of \textit{randomness factors} that are considered in the investigation significantly increases the computation cost \cite{bouthillier_accounting_2021}. To overcome this problem, more sophisticated experiments should be designed.

Focusing on multiple randomness factors at the same time, across multiple approaches is only starting to appear, even though it allows us to draw more interesting conclusions and better focus on the most important factors~\cite{gundersen_reporting_2023, pecher2024fine, pecher2024sensitivity}.

\subsection{Limited Analysis and Explanation of Importance for the Effects of Randomness Factors}

A significant part missing when investigating the effects of randomness is a thorough analysis of the importance of the different \textit{randomness factors}. In many cases, the importance of \textit{randomness factor} may be obvious, such as when it causes a large variance in results ($> 10\%$). When the observed variance is smaller, a more detailed analysis and explanation is required. The effects measured as 1\% standard deviation have different significance when the performance is in high 90\% than when it is in low 20\%. However, almost all papers simply present a single aggregated value without further details or analysis.

An important missing analysis is the comparison of importance for different \textit{randomness factors}. As it is common to work with a limited computational budget in real-world scenarios (including a growing effort of green AI that aims to limit its  CO2 footprint), identifying the most important \textit{factors} would allow us to focus on them and help better allocate the resources for dealing with the effects of randomness. However only few papers compare between \textit{factors}, such as \cite{pham_problems_2021, zhao_closer_2021, jundi_how_2022, zhong_are_2021, zhao_calibrate_2021, bouthillier_accounting_2021, bouthillier_unreproducible_2019, pecher2024sensitivity}.

Finally, there is only limited analysis of how the sensitivity to the effects of different randomness factors changes across different approaches, datasets and models. Specific studies have already found that the importance of \textit{randomness factors} is not consistent across different approaches and modalities. Similarly, the analysis of how different systematic choices, such as the number of labelled samples or size of the dataset, affect the importance of randomness factors is largely missing, although some studies have started to focus on this problem.

\subsection{Disregarding Interactions Between Randomness Factors}
\label{sec:open-problems_disregarding-interactions}

Ignoring interactions between different \textit{randomness factors} can lead to biased results. We can observe that many papers often find contradictory findings across all the tasks for addressing the effects of randomness. The effects of interactions are most notable when evaluating the mitigation strategies, as this task has the least developed evaluation strategies, often using simple evaluation on a single \textit{randomness factor}. We argue that the interactions between \textit{randomness factors} are one of the primary contributors to the inconsistency in findings, as was already observed for \textit{in-context learning} (using optimised prompt format and high-quality samples making the sensitivity to data order disappear). 

The interactions should be taken into consideration in every task for addressing the effects of randomness, whether it is investigation of the effects, identifying the origin of randomness, proposing mitigation strategies, or even comparing and benchmarking different \textit{randomness factors} and areas of machine learning.  

One possibility is to include all randomness factors in the process, such as exploring through all the combinations of the \textit{randomness factor configurations} for each \textit{randomness factor}. However, this incurs significant computational cost due to the number of runs growing exponentially with the number of \textit{randomness factors} and so may be infeasible. Some papers already do this, but only for a set of selected \textit{randomness factors} with low number of \textit{randomness factor configurations} \cite{zhao_calibrate_2021, jundi_how_2022, zhao_closer_2021}.

Another possibility is to mitigate the effects of \textit{randomness factors} that are not currently investigated using mitigation strategies. However, such approaches are only starting to appear~\cite{pecher2024sensitivity, webson-pavlick-2022-prompt}.

Taking interactions into consideration also has an impact on the evaluation of the effects of randomness, estimation of the mitigation strategy effectiveness and comparisons between approaches and \textit{randomness factors}. Each such tasks need to be explicitly modified to take the interactions into consideration. Specifically, the effects of different \textit{randomness factors} need to be disentangled. Statistical approaches, designed in some papers, that can disentangle these effects already show promising results for dealing with this problem \cite{bouthillier_accounting_2021, dodge_fine-tuning_2020}.

\subsection{Oversimplified Techniques and Metrics Employed to Evaluate the Effects of Randomness}
\label{sec:open-problems_oversimplified-techniques-and-metrics}

The majority of the papers evaluate the experiments when addressing the effects of randomness in a basic manner, often using a single aggregated value based on the model prediction. As the number of runs is chosen arbitrarily, the estimation for the real distribution of the results may be skewed. Therefore, the evaluation is open to the still present effect of randomness which can lead to biased results.

More sophisticated evaluation strategies (based on statistical approaches) that can account for the still present uncertainty in the limited runs, or the ones that can better estimate the real distribution of results without the need for a large number of samples (training runs), need to be designed to deal with this problem. Currently, only a few works determine the effects of randomness based on such statistical approaches \cite{bouthillier_accounting_2021, dodge_fine-tuning_2020, sellam_multiberts_2022}. At the same time, these evaluation strategies need to be used throughout all the tasks for addressing the effects of randomness. Even though some more sophisticated approaches already exist, many tasks (such as evaluating the effectiveness of mitigation strategies) still use a basic evaluation based on aggregated value, often from a single run.

Another avenue for improvement is to design metrics that evaluate the effects of randomness based on the learned representation instead of using the metric that observes changes in model prediction and use them as surrogate for estimating the effects.

\subsection{Inconsistency in Findings from Different Tasks for Dealing with Effects of Randomness}

Our in-depth analysis of this survey revealed that there are many contradictory findings across papers. Even though the effects of randomness are mostly investigated using a consistent approach (i.e., running multiple training and evaluation runs), there are aspects of the investigation methodology that differ across papers. In some cases, it is observed that the \textit{randomness factors} behave in a similar fashion across datasets, modalities and architectures, while in other cases their behaviour is inconsistent~\cite{zhao2023survey, biderman2024lessons, atreja2024prompt}. 

We believe that these inconsistencies are partially a result of disregarding the interactions between \textit{randomness factors} (Section \ref{sec:open-problems_disregarding-interactions}) as well as using oversimplified techniques and metrics to evaluate effects of randomness (Section \ref{sec:open-problems_oversimplified-techniques-and-metrics}). As such, investigation strategies that can handle the interactions or statistical methods that can better disentangle and estimate the true effects should be used.

In addition, observed inconsistencies may be a result of insufficient evaluation across a low number of samples and runs; or a result of systematic choices. Following, we describe in more detail such sub-optimal and inconsistent decisions in the investigation and evaluation methodologies and how to properly address them to achieve more consistent findings:
\begin{itemize}
    \item \textbf{Low number of test samples and investigation runs.} The effects of randomness are often evaluated using a low number of test samples (100 - 1000) and a low number of repeated runs (10). This often leads to biased results due to the non-representative distribution of the results that is affected by randomness and limits reproducibility. Following the long-used practices in machine learning, both the number of test samples and repeated runs should be significantly higher, especially when dealing with limited labelled data that only strengthens the problems with randomness.
    \item \textbf{Systematic choices.} Although the experimental setup plays a significant role, it is the most inconsistent aspect across studies, such as using a different number of examples, models, model sizes, or hyperparameter setup. As such, the findings cannot be easily compared with each other. Many solutions for this exist, such as comprehensive reporting for the experimental setup, keeping the setup as similar as possible to previous studies, or designing and using an investigation framework that handles this.
\end{itemize}

Following the analogy of benchmarks, we imagine that many different investigation frameworks can be proposed, with their designs reflecting some specific aspect deemed to be important (e.g., handling interactions, comparing effects of different \textit{randomness factors}, computational cost). However, so far no such explicit, fully fleshed out framework exists, only starting points of ones, mainly in papers introducing better evaluation strategies \cite{boquet_reproducibility_2019, bouthillier_accounting_2021, dodge_fine-tuning_2020, sellam_multiberts_2022}.

\subsection{Absence of Effective and Efficient Mitigation Strategies for Multiple Randomness Factors}

Many partial solutions for mitigating the effects of different \textit{randomness factors} exist. Such mitigation strategies are explicitly designed to deal with specific origin of randomness and are always used one at a time. However, many \textit{randomness factors} have no specific mitigation strategy designed for them. In some cases, a specific mitigation strategy may not even be possible, such as when dealing with non-deterministic implementation and hardware.

The only mitigation strategy that is usable for all \textit{randomness factors} is the \textit{ensemble} and, in some cases, further fine-tuning of the models. Although these strategies can effectively mitigate the effects of randomness, they incur heavy computational costs, which makes them infeasible in combination with complex models. In addition, they often disregard interactions between randomness factors.

Mitigation strategies have the potential to have the highest impact on all tasks for dealing with the effects of randomness and on the use of learning with limited labelled data in practice. Having an effective and efficient mitigation strategy that can take the interactions into consideration will positively impact all open problems mentioned in this section. Therefore, it is important to focus on designing a mitigation strategy that can address the problem as a whole. 
The focus should be on effectively combining different \textit{randomness factor} specific mitigation strategies. Besides taking interactions into consideration and achieving a significant reduction in randomness, a specific focus should also be given to reducing the additional computation cost that is incurred. This will be especially important for the \textit{ensemble} strategy, as we see it being used for \textit{randomness factors} without specific strategies designed for them. Some papers, also from related domains, focus on reducing the cost of the ensemble strategy~\cite{pecher2024fighting, liang-etal-2022-camero, chang-etal-2023-multi, summers_nondeterminism_2021}.

\subsection{Limited Consideration For Randomness in Comparisons and Benchmarks}

Even though many modifications are proposed that can deal with the effects of randomness, they are applied only in specific cases of comparisons between approaches, but not as a part of tasks for addressing the effects of randomness. The results from experiments investigating the \textit{randomness factors} and especially determining the effectiveness of mitigation strategies are compared in a simple manner based only on aggregated value. When comparing mitigation strategies, it is common to even use a single run, which may obscure their benefit~\cite{pecher2024automatic, pecher2024fighting}. Similarly to comparing approaches, the experiments investigating randomness are still sensitive to the effects of randomness as not all \textit{randomness factors} are considered. More sophisticated comparison strategies should be applied when comparing between different \textit{randomness factors}.

In addition, the benchmarks are only starting to consider randomness in their design, even though they are significantly affected by the randomness. Even when randomness is considered in the design, it is usually in a simple manner, such as looking at simple characteristics, providing multiple splits to train on, using different prompts, or repeating the evaluation multiple times with small changes~\cite{zheng_fewnlu_2022, mukherjee_clues_2021, alzahrani2024benchmarks, madaan2024quantifying}. Even this little consideration is a step in a good direction, but we need benchmarks that explicitly take variance into consideration and use it to penalise approaches that are brittle in the presence of randomness, such as in \cite{efrat_lmentry_2022}. Finally, some \textit{randomness factors} are not even considered in benchmarks, such as model initialisation or the choice of data. Therefore, we still have a long way to go to have benchmarks that take into consideration all the important \textit{randomness factors}.

\section{Conclusion}
\label{sec:conclusion}

In this paper, we have filled the gap of a missing, comprehensive literature survey focused on the existing research works addressing the effects of randomness that negatively impact the stability in real-world settings when labelled data are lacking. We summarised the current approaches that address the effects of randomness by investigating these effects, determining their origin and designing mitigation strategies for different \textit{randomness factors}. In addition, we analysed a specific group of papers that aim to take randomness into consideration when comparing and benchmarking different machine-learning models. Based on the findings from the comprehensive analysis and synthesis of existing works, we identified a list of seven major challenges and open problems when dealing with the effects of randomness and provided some insights into how to handle them.

Although we focused on a comprehensive evaluation, the area of addressing randomness is currently quickly growing and has attracted a lot of attention recently, as many researchers recognised the sensitivity of different approaches to the effects of randomness. Since 2023, the number of papers addressing the sensitivity to the effects of randomness has almost tripled, with most of them focusing on in-context learning and mitigation. As such, we expect that new research works will be continuously released, the area will be evolving and hopefully the challenges and open problems will be addressed. 

This survey offers a long-term value as the identified challenges and open problems are integral to the area and have not been addressed sufficiently even with the recent increase in focus. We hope that this survey will help researchers more effectively understand the negative effects of randomness, the tasks performed when dealing with them, grasp its core challenges and better focus the attention to addressing the randomness and the open problems so that the field can be advanced. We expect that it will encourage researchers to focus on the underrepresented \textit{randomness factors}, their interactions, systematic choices and settings so that the effects of randomness and the behaviour of models when they are present can be understood in more detail and can be more easily taken into consideration and addressed (e.g., when designing benchmarks). Finally, we believe that this survey will allow future works to determine and compare how the area of addressing the sensitivity to the effects of randomness is continuously advancing.

\begin{acks}
This research was partially supported by TAILOR, a project funded by EU Horizon 2020 research and innovation programme under GA No. \href{https://doi.org/10.3030/952215}{952215}; DisAI, a project funded by European Union under the Horizon Europe, GA No. \href{https://doi.org/10.3030/101079164}{101079164}; and by vera.ai project funded by the European Union under the Horizon Europe, GA No. \href{https://doi.org/10.3030/101070093}{101070093}.
\end{acks}


\bibliographystyle{ACM-Reference-Format}
\bibliography{core,additional}

\appendix

\section{Detailed Paper Categorisation Using the Defined Taxonomy}
As a part of the main content of the survey, only a part of the full paper categorisation of the \textit{core} papers according to the defined taxonomy is included, mainly due to its size. The full paper categorisation, along with additional relevant information, such as a basic categorisation of the \textit{recognise} papers is a part of the digital appendix of this survey\footnote{Available at \url{https://kinit.sk/public/acm-csur-sensitivity-survey.html}}.

The following information is part of the digital appendix:
\begin{enumerate}
    \item \textbf{Full categorisation of \textit{core} papers} -- The categorisation is extended with additional taxonomy dimension as well as other metadata. The \textit{machine learning approaches} are further divided into more detail, e.g., in meta-learning, we identify which papers deal with optimisation-based meta-learning and which consider metric-based meta-learning. In addition, we also divide the \textit{tasks performed} to the full list presented in the survey to provide a better overview of which tasks the papers focus on when addressing the effects of randomness. We also add the \textit{dataset} secondary dimension that indicates what datasets the individual papers work with. Besides the changes to the taxonomy dimensions, we also include additional metadata, such as Document Object Identifier (DOI) number (where applicable), year of publication, or link to the paper. Finally, for majority of the papers, we also provide a short description of the main focus for the paper and its findings.
    \item \textbf{Basic categorisation of \textit{recognise papers}} -- We include the full list of the \textit{recognise} papers in this digital appendix, along with their metadata (authors, DOI, year of publication). In addition, we identify some of the \textit{randomness factors} the \textit{recognise} papers deal with. Finally, we specify a high-level idea for each paper, for example, whether the paper recognises the problem of instability, and tries to improve the situation.
    \item \textbf{Evolution in number of papers through years} -- Using the year of publication for each identified paper, we show the evolution in the number of papers that focus on addressing the effects of randomness.
    \item \textbf{Relation mapping between \textit{randomness factors} and \textit{machine learning approaches} for different \textit{tasks performed}} -- As a part of the survey, such mapping was presented in an aggregated form across all the tasks for addressing the effects of randomness. Besides the aggregated mapping, we also include the same table, but for each task separately. This mapping serves as an additional overview of what the individual papers focus on (\textit{randomness factors} in specific \textit{machine learning approaches} and \textit{tasks performed}).
\end{enumerate}

\section{Description of Machine Learning Approaches}
In this part, we provide a basic idea and high-level description of the different \textit{machine learning approaches} that specify our scope in the survey, along with some representatives for each. This should serve as a basic overview for understating the concepts of the approaches required for a better understanding of the surveyed papers, not as a comprehensive overview of the selected approaches. We provide description for \textit{meta-learning}, \textit{language model fine-tuning}, \textit{prompting/in-context learning}, \textit{prompt-based learning} and \textit{parameter-efficient fine-tuning} approaches.

\subsection{Meta-Learning}
The idea behind \textit{meta-learning} is to learn a general learning algorithm that can generalise across different tasks, with only a few samples available for each task. In essence, meta-learning learns a function (also called \textit{meta-learner}) that takes a set of training data as input and outputs the learned model parameters (also called \textit{base-learner}) that can be used for the specific task. The \textit{few-shot learning} setup is often used, where the samples are distributed across multiple tasks. During training, a subset of these tasks is sampled and split into training and testing set. Using the training tasks, the parameters of the \textit{base-learner} are optimised. The parameters of \textit{meta-learner} are then updated based on how good the \textit{base-learner} performs on tasks in the testing set. This training is performed over multiple iterations, while at the start of each iteration, the \textit{base-learner} is initialised or updated based on the information or knowledge contained in the \textit{meta-learner}. At test time, the \textit{base-learner} is primed for the specific task using a few adaptation (or support) samples. 

Different approaches of meta-learning differ based on the objective they optimise for and how the \textit{base-learner} and \textit{meta-learner} are defined. In \textit{optimisation-based} meta-learning, the objective is to find an optimal set of parameters in the \textit{meta-learner} that can be quickly adapted to any task when fine-tuned as a part of \textit{base-learner} on a given task. In optimisation learning, the \textit{base-learner} and \textit{meta-learner} often represent the same model, and the \textit{base-learner} is initialised by simply copying all the parameters of \textit{meta-learner}. The optimisation of the \textit{meta-learner} is performed similarly to the supervised learning (where the error is back-propagated through the whole network), but using a second-order derivative. Majority of optimisation-based approaches are based on Model-Agnostic Meta-Learning (MAML) \cite{finn2017maml}, as they are designed to overcome some of the problems present in MAML, for example reducing its computational complexity by approximating the outer loop (e.g., Reptile \cite{nichol2018reptile} or First Order MAML \cite{finn2017maml}) or optimising only part of the network while having the rest pre-trained (e.g. ANIL \cite{raghu2019anil}), or better dealing with task dissimilarity and distribution (e.g. MMAML \cite{vuorio2019mmaml}, LEOPARD \cite{bansal2020leopard}). 

In \textit{metric-based} meta-learning, the objective is to learn good representation for comparing adaptation samples with samples without known labels. The final prediction of the \textit{base-learner} is determined by the class of the most similar adaptation (or support) sample. The most popular approaches for metric-based meta-learning are Prototypical Networks \cite{snell2017prototypical}, or Matching Networks \cite{vinyals2016matching}, which only differ in how they function (e.g., comparing based on aggregated information from multiple samples or comparing between each sample separately). 

For a more comprehensive overview of the taxonomy of different \textit{meta-learning} approaches and a more detailed description, please refer to \cite{hospedales2021meta, lee2022meta, tian2022meta_learning_survey}.

\subsection{Language Model Fine-Tuning}

The main idea of \textit{language model fine-tuning} is to transfer knowledge from a pre-trained model to a specific task by updating its learned parameters using a few samples. The language model is first trained on large corpora of unrelated texts, or an already pre-trained language model is used. The last layer of such language model is then replaced, or a new one is added for the specific task. Afterwards, the language model is fine-tuned on the few samples available by simply running a training process for a few episodes on the samples with a much lower learning rate. 

As the process is pretty straightforward, the approaches differ only in what language model is used (which also defines what corpora it was pretrained on) and what part of the language model is fine-tuned. The common approaches include training all the layers, or only the last few layers of the language model (such as only the added classification layer), or even approaches that fine-tune specific, often disconnected layers \cite{lee2022surgical_finetuning}. The most popular language models that are used for fine-tuning are BERT \cite{kenton2019bert} and its modifications like RoBERTa \cite{liu2019roberta}, ALBERT \cite{lan2019albert}, DistilBERT \cite{sanh2019distilbert}, XLNet \cite{yang2019xlnet} and many others. 

For a more comprehensive overview of different \textit{language model fine-tuning} approaches, please refer to \cite{rogers2020primer, xia2020bert},

\subsection{Prompting and In-Context Learning}

Prompting (and its related technique in-context learning, also called few-shot prompting) is an emerging paradigm, where a large pre-trained language model is used for different tasks without first updating its parameters. Instead, all tasks are reformulated as a sequence generation problem and the language model is ``prompted'' to output a sequence of words. The resulting sequence of words is then mapped to a possible list of labels (e.g., the word ''good'' is mapped to positive and ''terrible'' to negative label in sentiment classification). To prime the model for a specific task, the presented input is constructed as a concatenation of instruction for the task, optionally a few labelled samples (serving as context) if in-context learning is used, and a single test sample, for which the label is predicted.

As the final prompt is a concatenation of multiple labelled samples and a single unlabelled one, usually only language models that allow large input sizes can be used. Another important choice in prompting and in-context learning is how the input prompt is designed, either manually, semi-automatically or automatically. The approaches differ on what language models are used, how the prompt format is designed and whether the prompting and in-context learning is combined with some kind of fine-tuning. The most popular model used in in-context learning is GPT-3, with the approaches starting to design automatically generated prompts, although also smaller models are used when combined with fine-tuning \cite{brown2020language, gao2021making, perez2021true, logan2022cutting}.

For a more comprehensive overview of different \textit{in-context learning} approaches, please refer to \cite{liu2021pre, dong2022survey, song2023comprehensive_survey_of_fsl}.

\subsection{Prompt-Based Learning}

Prompt-based learning, often also called instruction-tuning, can be viewed as a modification of typical fine-tuning for the large generative language models and their use through prompting and in-context. The idea is to bridge the gap between the next-word prediction objective of the large language models and the objective of the model adhering to human instructions. As such, the goal is to optimise the parameters of the large language models to better follow the instructions that are included as part of the prompts.

Following the typical fine-tuning technique, a pre-trained large language model is further trained on a dataset consisting of pairs of instructions and the outputs, where the instructions denote the instruction in the input prompt and the outputs represent the generated words that are mapped to classes. Such tuning allows for more controllable and predictable model behaviour, improving the capability of the models to perform the specific task by improving the mapping from instructions to the generated words that are further mapped to classes. In addition, it is characterised by the benefits of fine-tuning, i.e., the rapid adaptation to a specific domain and task without extensive retraining or architectural changes.

Even though it provides its benefits, it also faces challenges, especially when it comes to the sensitivity to the effects of randomness. As we are combining the fine-tuning and prompting/in-context learning techniques, it also combines the sensitivity of these models to the randomness factors in fine-tuning and in-context learning. For example, designing high-quality prompts or instructions, the choice of samples, their order, but also the whole optimisation process.

For a more comprehensive overview of different \textit{prompt-based learning} approaches, please refer to~\cite{zhang2023instruction, li2024visionlanguage}.

\subsection{Parameter-Efficient Fine-Tuning}

Parameter-Efficient Fine-Tuning (PEFT) is an extension of regular fine-tuning, where the optimisation of the parameters in the pre-trained large models is done whoile minimising the number of additional parameters introduced or the computational resources we require. This is particularly important for the massive generative language models with high parameter counts, such as GPT-3, that are often used through prompting or in-context learning. Although PEFT methods are more popular for such models, they can be utilised for any kind of model to reduce the required resources to run them. 

Overall, there are a few categories of PEFT methods, based on how they reduce the number of trainable parameters. First, additive methods, keep all the parameters of the pre-trained models fixed and only add some additional parameters that are trained and combined with the frozen parameters. This includes approaches such as the use of Adapters (e.g., Pfeiffer Adapter)~\cite{houlsby2019parameter, pfeiffer-etal-2021-adapterfusion}, or optimising soft-prompt (e.g., prefix-tuning~\cite{li-liang-2021-prefix}, p-tuning~\cite{liu2023gpt} or prompt-tuning~\cite{lester-etal-2021-power}). Second, selective methods, do not introduce additional parameters, but instead choose a subset of the pre-trained model parameters to train. This includes approaches based on lottery ticket hypothesis~\cite{ansell2021composable} or other approaches for choosing the subsets~\cite{xu_raise_2021, fu2023effectiveness}. Third, reparametrised approaches, introduce additional low-rank trainable parameters during training that are then fused with the original model for inference. The most popular approach is the low-rank adaptation (LoRA)~\cite{hu2022lora} and its alternatives. Finally, there are also hybrid approaches that combine multiple PEFT methods.

For a more comprehensive overview of different \textit{parameter-efficient fine-tuning} approaches, please refer to~\cite{han2024parameter}.

\section{Implementation of the Survey Methodology}
\label{sec:appendix_methodology}
\subsection{Search query definition}

The search query is composed of two parts: 1) \textit{terms} that represent all the versions of stability that we consider in this survey; and 2) \textit{scope} describing our focus on approaches that can be used for learning with limited labelled data.

The set of terms that can be used to describe the effects of randomness on the performance was progressively expanded based on the terms in papers that in any way address this randomness. The identification of the papers was done randomly by first sampling papers that reference \textit{stability}, assessing whether their understanding of stability adheres to our definition, and using the cited and citing papers to identify the next potential candidates. Using this methodology we identified the first set of key-words that included the following words: \textit{stability}, \textit{instability}, \textit{sensitivity}, \textit{variance}, \textit{randomness}, \textit{robustness} and \textit{reproducibility}. However, we decided to remove the terms \textit{robustness} and \textit{reproducibility} from this list, as they have much wider meaning and would lead to many irrelevant papers.

The \textit{scope} was defined based on the groups of approaches we want to focus on. Based on reading the papers, we have noticed only two terms being widely used to define all the approaches we were interested in: \textit{meta-learning} and \textit{few-shot learning}. We decided to use the term \textit{few-shot learning} instead of \textit{learning with limited labelled data}, because of its popularity and its misuse leading to both terms meaning the same thing in the literature. We noticed many different uses of these terms across papers, with many removing the hyphenation or foregoing the "learning" part of \textit{few-shot learning}. To account for all the different uses of the terms, we finally decided to use the most simplified version of them, specifically \textit{meta learning}, \textit{metalearning} and \textit{few shot}.

The final query we used for the keyword search was: `(``few shot'' OR ``meta learning'' OR ``metalearning'') AND (``stability'' OR ``instability'' OR ``sensitivity'' OR ``variance'' OR ``randomness'')`. We also searched only for papers that were published in the year 2017 or later, as the focus on the effects of randomness and stability before this year was non-existent.

\subsection{Relevant Digital Libraries Used to Discover Papers}

We used digital libraries that provide flexible search options, with a focus on the machine learning domain as one of its domains and allow for reproducible search results to discover the relevant papers. As we were not interested in other disciplines, we limited the search results to only those relevant for us. This includes the following databases: ACM Digital Library, IEEE Xplore and Scopus database.

In addition, we also searched through the proceedings of the top-rated machine learning conferences, such as ICML, ICLR, or NeurIPS. We have noticed that many relevant papers are submitted to these conferences, while at the same time only a fraction of them are indexed by the previously stated databases. We specifically searched through ACL Anthology (containing the aggregated proceedings of the ACL conferences), the NeurIPS conference and the Proceedings of Machine Learning Research (PMLR), which includes proceedings from conferences such as ICML or ICLR. Only the ACL anthology allows for advanced search through papers, nevertheless, the search is limited to only the top 100 results. Therefore, to search through these libraries we opted to use Google Scholar. Each conference was searched separately using a modified query to limit results to only specific site (using the keyword in search, i.e. `site:aclanthology.org` for ACL, `site:neurips.cc` for NeurIPS and `site:mlr.press` for PMLR).  We take only the top 200 relevant papers from each conference. To determine the cut-off point, we manually determined the relevance of the papers at later pages of results and noticed the relevance drops of significantly after $\sim$100 papers. Therefore we decided to search through 100 more papers just to be sure we were not missing any relevant papers.

Lastly, we also use included cited and citing papers from the most relevant ones as additional sources. We detected a significant clustering behaviour in papers that deal with the effects of randomness, with papers related to specific groups of approaches extensively citing all other papers in this group. This significantly improves the coverage of results as it includes papers not indexed in the databases, but also those that may use slightly different terminology or those that are relevant, but do not focus only on the limited data setting.

\subsection{Identification of relevant papers using search and filtering}

Since the research paper collection from digital libraries would naturally result in a large set of papers with various relevance, we selected the subset of the most relevant papers that we call \textit{core} papers. To identify the set of these papers, we perform the following steps across multiple iterations:
\begin{enumerate}
    \item \textbf{Create a set of potentially relevant papers.} The defined search query is used across all specified digital libraries to create the set of papers that are potentially relevant for the work. This generates a large set of papers with many false positives.
    \item \textbf{Filter papers based on relevance.} The large set of potential papers is reduced to a set of papers dealing to some extent with the effects of randomness. To filter out the irrelevant papers the context in which the individual key-words from the search query are used in the paper. This is done by finding all the appearances of the key-words using a full-text search and manually checking them. After finding the first relevant context for both the \textit{term} and \textit{scope} key-words, the paper is considered to be relevant. If no such context exists for one of the groups, or the context indicates a different definition of the problem (such as \textit{algorithmic stability}), the paper is considered to be irrelevant and removed from the set. For example, we do not consider the context to be relevant if it appears in related work. This step significantly reduces the set, leaving only \~5\% of papers.
    \item \textbf{Filter papers based on merit.} Although the papers in this step are mostly relevant, the set still contains papers we do not want to consider based on their form. This includes papers that are only extended abstracts, proposals for talks or projects, or papers appearing on papers not relevant to stability. In addition, papers from conferences with a lower Core rank than B or from journals not belonging to Q2 or Q1 are also removed.
    \item \textbf{Filter papers based on focus.} In this step we filter out the papers that only recognise that there is a problem with stability in regards to randomness in some approaches but provide no further investigation or focus on the randomness. We call these papers as \textit{recognise} papers. This mostly includes papers that note that the training process with limited data is unstable so they perform multiple runs, or papers that note that it was found in previous papers that there is a problem. This leaves us with a set of papers we consider to be \textit{core}.
    \item \textbf{Extend the set of papers with citing and cited papers.} As there is a strong clustering present, papers dealing with similar approaches cite each other extensively. We make use of this clustering to discover other relevant papers that may not be indexed by the databases or that use different terminology so the search query does not catch them. However, we extract such papers only based on the context they are referenced in (when cited in the paper), or based on the relevant filtering step applied only to the title and abstract of the paper (which cites the current paper). We identify the citing papers through the Google Scholar \textit{cited by} section present next to each paper.
\end{enumerate}

We repeat the steps $2-5$ until no new papers are identified in step $5$.

\subsection{Analysis and categorisation of the papers}

All the papers identified this way were then manually categorised using our taxonomy as described in the main content of the survey (Section 3). The content of each paper was analysed and assigned value for each defined property. Afterwards, the main categorisation property (task performed to address the randomness) was selected based on how best it splits the papers into independent groups.

For each task for addressing the effects of randomness, additional properties and characteristics were identified and used to cluster the papers further. These additional properties were then used to derive the main findings from each task. Therefore, each finding can be mapped to one or multiple such properties at the same time. Besides the main properties and characteristics described in Section 3, we consider also the following additional properties and characteristics of the papers:
\begin{enumerate}
    \item \textit{Modality} that divides the papers into those focusing on text data, images, tabular data, or any combination of these modalities.
    \item \textit{Groups of approaches} further divide the primary property of the \textit{machine learning approach} into a more detailed categorisation. For example, the \textit{meta-learning} approaches are further divided into optimisation or metric-based meta-learning.
    \item \textit{Interactions}, a binary property that specifies whether the confounding effects between different \textit{randomness factors} are addressed in any way in the paper.
    \item \textit{Standard deviation}, a binary property that specifies whether the performance is reported using standard deviation (or similar metric such as confidence intervals) or just as a simple aggregate metric (mean, median, etc.).
    \item \textit{Out-of-distribution}, a binary property that specifies whether the data used for testing purposes in the papers comes from the same distribution as the training data, such as addressing the effects of randomness when dealing with multilingual data or data drifts.
    \item \textit{Datasets} used to better explore what datasets are popular when addressing the effects of randomness.
\end{enumerate}

The challenges and open problems were identified by observing common patterns in the findings, such as contradictions in results. In addition, the open problems were further extended by identifying parts of the research process that are currently missing even though their inclusion is a straightforward and simple modification of the current approach, such as using statistical evaluation of results instead of reporting a single performance metric.

\end{document}